\documentclass[a4paper,fleqn]{cas-dc}

\usepackage[numbers]{natbib}
\usepackage{rotating}
\usepackage{bbding}
\usepackage{subfigure}
\usepackage{booktabs}
\def\tsc#1{\csdef{#1}{\textsc{\lowercase{#1}}\xspace}}
\tsc{WGM}
\tsc{QE}
\tsc{EP}
\tsc{PMS}
\tsc{BEC}
\tsc{DE}

\begin{document}
\let\WriteBookmarks\relax
\def\floatpagepagefraction{1}
\def\textpagefraction{.001}

\shorttitle{Knowledge Based Systems}

\shortauthors{H. Tu, Z. Yang, J. Yang et~al.}

\title [mode = title]{PCAE: A Framework of Plug-in Conditional Auto-Encoder for Controllable Text Generation}                      



%






\author[1]{Haoqin~Tu}
\author[1]{Zhongliang~Yang}
\ead{yangzl15@tsinghua.org.cn}

\author[1]{Jinshuai~Yang}



\author%
[1]
{Siyu~Zhang}


\author%
[1]
{Yongfeng~Huang}
\address[1]{Department of Electronic Engineering, Tsinghua University,
    Beijing,
    100084, 
    China}



\begin{abstract}
Controllable text generation has taken a gigantic step forward these days. Yet existing methods are either constrained in a one-off pattern or not efficient enough for receiving multiple conditions at every generation stage. We propose a model-agnostic framework \textbf{P}lug-in \textbf{C}onditional \textbf{A}uto-\textbf{E}ncoder for Controllable Text Generation (\textbf{PCAE}) towards flexible and semi-supervised text generation. Our framework is ``plug-and-play'' with partial parameters to be fine-tuned in the pre-trained model (less than a half). Crucial to the success of PCAE is the proposed broadcasting label fusion network for navigating the global latent code to a specified local and confined space. Visualization of the local latent prior well confirms the primary devotion in hidden space of the proposed model. Moreover, extensive experiments across five related generation tasks (from 2 conditions up to 10 conditions) on both RNN-based and pre-trained BART \cite{lewis2020bart} based auto-encoders reveal the high capability of PCAE, which enables generation that is highly manipulable, syntactically diverse and time-saving with minimum labeled samples. We will release our code in \url{https://github.com/ImKeTT/pcae}.
\end{abstract}




\maketitle
\section{Introduction}
Obtaining systems to automatically produce realistic-looking texts has been a goal pursued since the early stage of artificial intelligence \cite{meehan1977tale}. In real life scenarios, to approach more human-like contexts, the generated sentences should be tailored to their specific audience \cite{garbacea2020neural}. As a result, controllable text generation (CTG) has drawn great attention nowadays \cite{ficler2017controlling,hu2017toward,dathathri2019plug}. Controllable text generation aims at generating coherent and grammatically correct texts whose attributes can be controlled \cite{elhadad1990constraint}, and/or abide by user-defined rules which reflect the particular interests of system users \cite{garbacea2020neural}.

\begin{figure}[]
\centering
\includegraphics[width=\linewidth]{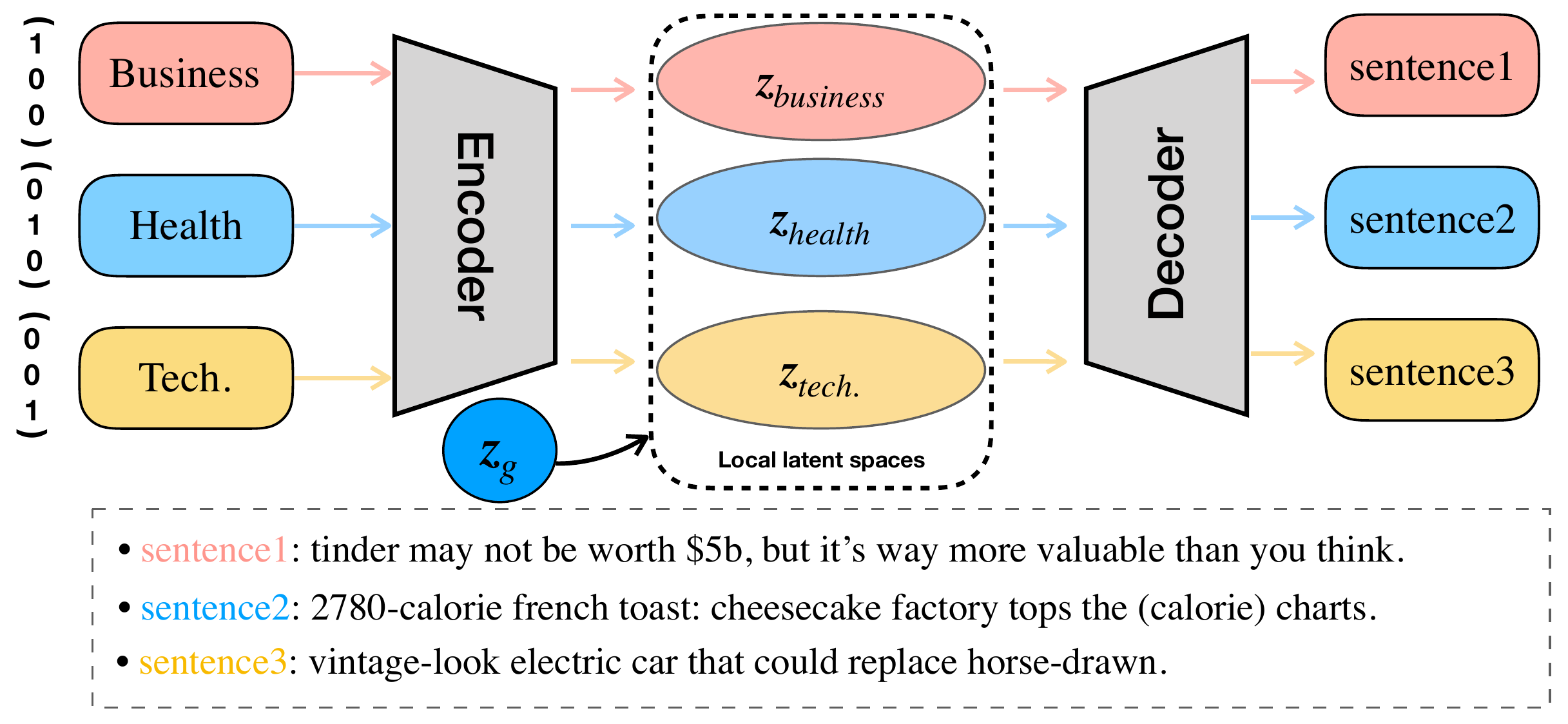}
\caption{A running example of the CTG task using auto-encoders. For controllable generation, we only need to input control signals (i.e., one-hot class label) and a global latent vector $\boldsymbol{z_g}$ sampled from standard Gaussian. Then the model produces texts that fulfill given conditions by creating specified local latent spaces.}
\label{fig:task}
\end{figure}

With the successful deployment of deep neural networks, recent proposed methods have brought us closer to this objective by producing texts with specified attributes. A general idea is to embed given conditions into an end-to-end training scheme \cite{kingma2014semi,hu2017toward,li2020optimus} in order to produce sentences that fulfill given conditions, which has been illustrated in Figure \ref{fig:task}. Nevertheless, there are two main defects of these models that limit the application of these methods in reality. Firstly, these methods cannot deal well with real-world cases where conditions are not distributed at one time, i.e., new conditions for new using circumstance. In this scenario, models like SVAE \cite{kingma2014semi}, OPTIMUS \cite{li2020optimus} need to activate all model parameters to be trained for these new conditions, which are time-wasting, thus are not the ideal re-deployments for practical use \cite{houlsby2019parameter}. 
Secondly, these models are mostly restricted to custom and well-designed language models, which means it is inconvenient to apply them directly to other more advanced language models for better modeling results. To address these problems for more practical application, another line for CTG follows the Pre-train and Plug-in (PnP) paradigm \cite{dathathri2019plug} has raised a lot research focus in recent years. By freezing the base language model (LM) and modifying few or no plug-in parameters, this paradigm is more flexible and powerful for controllable generation since it is parameter-efficient and can be applied to any advanced LM. Despite its success, there are two main defects regard to existing PnP works: one is that they are not convenient for creating texts with numerous categories at one time. Take currently the best-performed PnP language model PPVAE \cite{duan2020pre} as an example, when $n$ conditions come in at some point, it demands to train additional $n$ plug-in AEs to produce controlled texts. This drawback makes the whole system verbose and time-wasting when it meets a great amount of conditions. Another issue is that these PnP methods with only hidden mapping functions to be updated during plug-in process may be incapable of reaching a high degree of control.
Since auto-encoders have shown favourable learning ability of text integral properties that are beneficial for controllable generation \cite{fang2019implicit}, we extend from existing text AE-based \cite{bowman2016generating} PnP frameworks, and isolate the textual syntax module from the input condition representation module by building the \textbf{BaseAE} and \textbf{PluginAE} separately.

Formally, the \textbf{BaseAE} can be any kind of text auto-encoder, which takes the main responsibility to formulize the basic sentence generation guidance as a standard LM. To benefit PluginAE in its high-dimensional hidden space, BaseAE is also expected to expatiate a robust and continuous latent manifold. As for \textbf{PluginAE}, it is a model-agnostic lightweight inserted component for BaseAE. Our PluginAE architecture is designed with efficient broadcasting label infuser \textit{Broadcast Net}, incorporating label prior to BaseAE's latent space and enabling the plug-in model to learn all the conditions with one single training procedure. In purpose to achieve a higher level of control for our model, we choose to activate the decoder originated from the BaseAE during plug-in training. Our contributions can be listed as follow: 
\begin{enumerate}
    \item We explored a novel model-agnostic controllable text generation method PCAE. It is based on PnP framework and can be easily adopted to any kind of advanced auto-encoders for text controllable generation.
    \item We devised the \textit{Broadcast Net} for efficient fusion between conditions (labels) and latent space, so the model can generate controllable texts with very few labeled samples and time.
    \item To explain the advantageous improvement of PCAE, we evaluated our model on five different related tasks with conditions ranging from 2 to 10. We further utilized both RNN-based and pre-trained BART \cite{lewis2020bart} based auto-encoder to verify the effectiveness of proposed framework.
\end{enumerate}
Inspiring results demonstrate that our model is both time-saving (reduce up to $35\%$) and highly controllable (near $90\%$ accuracy with $100$ labels for each class in the best case) compared with both competent RNN-based and BART-based baseline language models.

\section{Related Work}
\subsection{Text Auto-Encoders with Latent Variables}
Latent variable models (LVM) have drawn massive attention in text generation field \cite{bowman2016generating,wang2019topic,tang2019topic,duan2020pre}. The latent space geometry of LVMs can conduct multiple view of knowledge in a given corpus (i.e., style, topic, and high-level linguistic or semantic features). There are two famous categories for text modeling with auto-encoders (AE), namely variational auto-encoders (VAE) \cite{bowman2016generating} and adversarial auto-encoders (AAE) \cite{makhzani2015adversarial}. They commonly employ the evidence lower bound (ELBO) maximization of data $p_{\theta}(\boldsymbol{X})$ to update the holistic model. A major distinct between these two models lies in the regularization term of their ELBOs. While VAE takes a Kullback-Leibler (KL) penalty as its latent regulator, AAE introduces a discriminator to judge latent differences as illustrated below, 
\begin{equation}
    \begin{aligned}
    &\log p(\boldsymbol{X}) \geq \underbrace{\mathbb{E}_{q(\boldsymbol{z} \mid \boldsymbol{X})}[\log p(\boldsymbol{X} \mid \boldsymbol{z})]}_{\text{reconstruction term}}\\
    &-\left\{\begin{aligned}
    &\underbrace{\mathbb{D}_{\mathrm{KL}}(q(\boldsymbol{z} \mid \boldsymbol{X}) \| p(\boldsymbol{z}))}_{\text{KL penalty}}\quad \text{ELBO of VAE}\\
    &\mathbb{E}_{p(z)}[-\log D(\boldsymbol{z})]\quad \text{ELBO of AAE}\\&\underbrace{+\mathbb{E}_{p(\boldsymbol{X})}[-\log (1-D(E(\boldsymbol{X})))]}_{\text{Discriminator penalty}}
    \end{aligned}\right.,
    \end{aligned}
    \label{eq0}
\end{equation}
where function $D(\cdot)$ and $E(\cdot)$ for the ELBO of AAE denote its discriminator and encoder respectively. The VAE as a general tool is widely used in continuous generation (e.g., image generation). However, when it comes to the discrete domain (i.e., text generation), VAE is facing numerous plights, such as latent vacancy dilemma \cite{xu2020variational}, latent vanishing problem \cite{bowman2016generating}, etc. The main reason is that VAE often neglects latent information provided by the encoder. In contrast to VAEs, AAEs maintain a strong coupling between their encoder and decoder, ensuring that the decoder does not ignore representations in the latent space, which makes it robust for latent knowledge interpretation and interpolation \cite{vincent2010stacked,makhzani2015adversarial}. However, \citet{li2020optimus} proved that a strong encoder such as pre-trained BERT in a VAE is very helpful to remit such issue. As a result, we employed AAE loss for RNN-based PCAE and VAE loss for pre-trained BART-based PCAE to show our framework is model-agnostic and effective under any auto-encoder.


\subsection{Auto-Encoders with Pre-trained Language Models}
Large pre-trained language models (PLMs) are gaining more and more popularity these days. With enormous resources being devoted, experienced encoders\&decoders such as BERT \cite{devlin2019bert}, GPT-2 \cite{radford2019language} and T5 \cite{raffel2020exploring} are devised to fully understand textual contents and create human-like sentences respectively. Incorporating these mighty PLMs as encoder and decoder of a variational auto-encoder can largely mitigate the KL collapse problem by offering the decoder a nonnegligible latent space from its encoder \cite{li2020optimus}. Several works to incoporate these PLMs into latent auto-encoders have been explored nowadays \cite{liu2019transformer,li2020optimus,fang2021transformer,park2021finetuning,tu2022adavae}, which have shown promising potential in a varied multitude of tasks including unsupervised latent interpolation \cite{li2020optimus,park2021finetuning}, controllable text generation \cite{li2020optimus} and prompt story generation \cite{fang2021transformer}, etc.

\subsection{Controllable Text Generation} 
The core idea of controllable text generation is to generate textual contents with designated conditions to cope with specified circumstances and auditors. Formally, we follow the problem setting in previous works \cite{hu2017toward,duan2020pre} to define the task: Given a set of $k$ conditions $\boldsymbol{L}=\{l_1, l_2,...,l_k\}$ (e.g., specific topics, sentiment labels), conditional text data $\boldsymbol{Y}=\{\boldsymbol{Y}_1, \boldsymbol{Y}_2,...,\boldsymbol{Y}_k\}$ and unlabeled corpus $\boldsymbol{X}$, where each text corpus $\boldsymbol{Y}_i$ corresponds to its label $l_i$. With condition label $l_i$ as input, we aim at learning a language model $\mathcal{F}(l_i)$ to calculate the distribution over the text samples $\boldsymbol{Y}_i$. Thus, when the condition $l_i$ is specified, the model could generate realistic text samples that fulfill the given condition. And in practice, we usually leverage a trained text classifier to distinguish texts with different concepts (see Sec. \ref{sec:control} for controllability analysis).

To support generating sentences that fulfill such request, recent researches are mainly divided into threefold according to their training paradigm: supervised, self-supervised and semi-supervised. For fully supervised methods, adversarial components like specified discriminators are widely employed \cite{chen2016infogan,wang2018sentigan}. In spite of their high controllability, they require abundant labeled data and enormous computational resources, which is unpractical for real world applications. Self-supervised methods commonly explore the hidden embeddings of LMs \cite{wang2018sentigan,tang2019topic} and apply themselves to catch the underlying control rules during training, yet they normally provide sequences with a low degree of control. 

The third party is semi-supervised, which requires only limited labeled data for controllable generation. SVAE \cite{kingma2014semi} as the first semi-supervised VAE model, was initially applied to visual domain. \citet{duan2020pre} explored its modeling formulation into language domain, which treats the label embedding as an extended part of the latent variable when there are label-text pairs available. \citet{li2020optimus} proposed OPTIMUS with BERT and GPT-2 as encoder and decoder respectively. They conducted controllable text generation via a latent space adversarial network using a two-stage training, which only requires labeled data at the second stage. 

Apart from SVAE and OPTIMUS, one important branch named ``Pre-train and Plug-in'' (also known as plug-and-play) is rising recently. Since labeled samples are generally required only at ``Plug-in'' stage in PnP models, their training fashion is categorized as semi-supervised. \citet{keskar2019ctrl} used human-defined ``control code'' to pre-trained LMs in order to generate controllable texts, but needs full-scale fine-tuning. To reduce training time, \cite{dathathri2019plug} firstly proposed the concept of plug-and-play for conditional text generation, which generates controlled sentences by pulling the gradients of LMs along the desired path using extra components with few parameters. However, it was proposed based on large pre-trained language models and still requires hours to be trained. What followed was the PPVAE \cite{duan2020pre}, which can be inserted to any pre-trained AE to create conditional texts. Nevertheless, it does not equip label infuser to incorporate condition knowledge explicitly into generation, thus has to train $n$ plug-in VAEs when $n$ new conditions come in. To focus on a fine-grained generation, \citet{mai2020plug} further extended the paradigm of PnP to text style transfer, which treats target texts as labels and employs a novel ``offset'' net as well as the latent adversarial loss for generation. Other lines of PnP controllable generation either targets at changing the prompts/prefix to be fed into the base LMs during training procedure \cite{wallace2019universal,li2021prefix}, or shifting output probabilities from trained LMs at inference time \cite{krause2021gedi,pascual2020directed}. These methods are mostly based on large pre-trained models and generally take hours to be fully tamed (sometimes their training times are even longer than fine-tuning) \cite{krause2021gedi,li2021prefix,he2021towards}.

\begin{table}
\centering
\begin{tabular}{cl} 
\toprule[1.5pt]
\textbf{Variable}                & \multicolumn{1}{c}{\textbf{Description}}                                                 \\ 
\hline
$\boldsymbol{X}$        & Input unlabeled text corpus                                 \\
$\boldsymbol{Y}$        & Input labeled text corpus                                   \\
$y_i$                   & The $i$-th word from a data point in~$\boldsymbol{Y}$        \\
$\boldsymbol{L}$        & Task label set                                              \\
$l_i$                   & The $i$-th label from the label set                         \\
$\boldsymbol{Y_i}$      & Labeled text corpus with label $l_i$                        \\
$\boldsymbol{Z_g}$      & Global latent space                                         \\
$\boldsymbol{z_g}$      & Global latent vector from $\boldsymbol{Z_g}$                \\
$\boldsymbol{Z_l}$      & Local latent space                                          \\
$\boldsymbol{z_l}$      & Local latent vector from $\boldsymbol{Z_l}$                 \\
$\mathcal{F}_l$         & Label embedding network                                     \\
$\boldsymbol{e_{l_i}}$  & Label embedding of label $l_i$                              \\
$\mathcal{F}_{z(t)}$    & The $t$-th latent transformation network                     \\
$\boldsymbol{z_{l(t)}}$ & The local latent vector after $\mathcal{F}_{z(t)}$          \\
$\boldsymbol{h_i}$      & The $i$-th hidden state of the decoder                       \\
$E(\cdot)$              & The encoder of models                                       \\
$D(\cdot)$              & The latent discriminator of AAE models                      \\
$k(\cdot,\cdot)$        & The kernel function                                         \\
$p(\cdot)$              & The prior distribution                                      \\
$q(\cdot)$              & The posterior distribution                                  \\
\bottomrule[1.5pt]
\end{tabular}
\caption{The main variable denotations in our method.}
\label{tab:variable}
\end{table}

\section{PCAE Methodology}
\begin{figure*}[ht]
\centering
\includegraphics[width=1.0\linewidth]{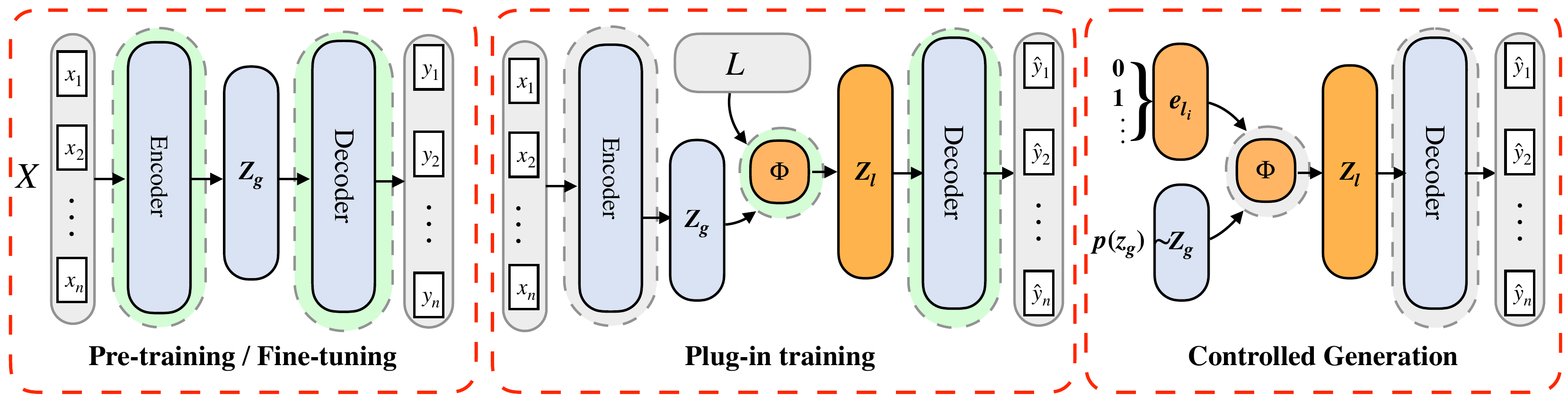}
\caption{A detailed workflow of the proposed framework. Parameters in components with green or gray backgrounds are activated or frozen respectively. During plug-in training, there is a label fusion function $\Phi$ (specifically the \textit{Broadcasting Net}) for a more effective controlled sentence inference. During generation, we produce controllable texts by assigning a desired class label to the trained PluginAE.}
\label{figmainmodel}
\end{figure*}

We present the main variable denotations in Table \ref{tab:variable}. The key idea of our framework is to reduce the resource consumption of training a language model with high controllability. The PnP framework with one full model training and plug-in controllable components is an efficient and flexible for this demand. Thus our model is separated into two disconnected sections: \textbf{BaseAE} and \textbf{PluginAE}, which corresponds to pre-training and plug-in training stage respectively. The model's workflow is in Figure \ref{figmainmodel}: the first figure represents the model structure of \textbf{BaseAE}, while the second figure is the structure of \textbf{PluginAE}. As for the third figure, it is the process for controllable text generation, which requires components from both \textbf{BaseAE} and \textbf{PluginAE}.

For pre-training stage, we use unlabeled textual data $\boldsymbol{X}$ to train the BaseAE language model (train from the scratch for RNN-based model and fine-tuning for BART-based model). For plug-in training, we input text-label pair $\{\boldsymbol{Y}, \boldsymbol{L}\} = \{\boldsymbol{Y}_i, l_i\}_i$, where $\boldsymbol{Y}_i$ is the training corpus from $\boldsymbol{Y}$ with label $l_i$. We use the labeled data pairs for conditional training in order to obtain the controllable decoder of PluginAE, which takes the latent variable and label condition $l_i$ to generate controllable texts. Thus, once we trained the PluginAE, we only need to input the sampled global latent vector from its prior $\boldsymbol{z_g}\sim N(0, I)$ and a control label $l_i$ (one-hot label) to the model for controlled generation. This training process makes PCAE only access to labels at the second stage, which makes it semi-supervised.
\subsection{\textbf{BaseAE} Pre-training/Fine-tuning}
BaseAE is under the obligation to present fluent and diverse texts and further derive meaningful latent representations. A BaseAE consists of three main components: the encoder, global latent space $\boldsymbol{Z_g}$ and the decoder. It should be noted that to ensure our BaseAE is a qualified LM, it ought to, in principle, be pre-trained on a very large text corpus (the unlabeled text data $\boldsymbol{X}$). As we employ both RNN and pre-trained BART in our framework, the pre-training stage of BaseAE represents training from the scratch for RNN and fine-tuning for BART-based model. For RNN-based BaseAE, we chose de-noising adversarial auto-encoder \cite{shen2020educating}, whose loss function is the same shown in Eq.(\ref{eq0}) except replacing $\boldsymbol{z}$ with global latent code $\boldsymbol{z_g}\sim \boldsymbol{Z_g}$. For BART-based BaseAE, we simply employed the BART encoder, decoder and the plain VAE training loss presented in Eq (\ref{eq0}) with $\boldsymbol{z}$ replaced with $\boldsymbol{z_g}$.
\begin{figure}[ht]
\centering
\includegraphics[width=0.80\linewidth]{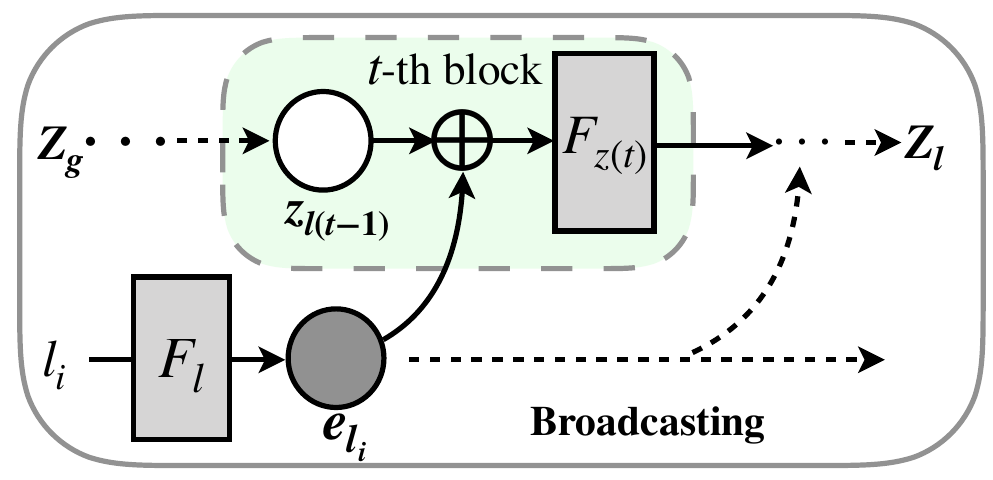}
\caption{The label fusion function $\Phi$, which is essentially a broadcast operation between label embedding and latent codes. Thus it refers to the \textit{Broadcasting Net}.}
\label{figlabel}
\end{figure}
\subsection{\textbf{PluginAE} Plug-in Training}
PluinAE is the BaseAE model with additional plug-in components. It is the direct portal for controllable text generation, which forms a more structured local latent space with label signals and global latent space from BaseAE for controlled generation. 
To make the PluginAE controllable for generation, we send global latent embedding and label signal to neural networks to produce a locally structured latent space $\boldsymbol{Z_l}$, and further feed it to the decoder for generation.

Inspired by DenseNet \cite{huang2017densely}, which densely connects neural representations with the latter layers of the network using skip connection \cite{he2016deep}, we utilize a label infuser that incorporates dense connection to sample the local variable $\boldsymbol{z_l}$ conditioned on both label and global latent information. That is to say, for given label representation, we broadcast it to every layer of a neural network using skip connection, which then produces the local latent vectors as output. We call this label infuser the \textit{Broadcasting Net}. Specifically, for a \textit{Broadcasting Net} with $n$ layers, it takes the previous latent vectors (i.e., global latent vector $\boldsymbol{z_g}$ at the beginning) and any label $l_i$ to generate the corresponding local latent vector $\boldsymbol{z_l}$, which can be formulized as: 
\begin{enumerate}
    \item For input one-hot label $l_i$, we obtain its representation $\boldsymbol{e_{l_i}}$ by label embedding layer $\mathcal{F}_l$.
    \item Given $\boldsymbol{z_g}$, we draw $\boldsymbol{z_l}$ by sampling from the local latent distribution $p(\boldsymbol{z_l}\mid \boldsymbol{z_g}, \boldsymbol{e_{l_i}})$. In detail, at $t$-th layer, this transformation with $\boldsymbol{z_g}$ and $\boldsymbol{e_{l_i}}$ is implemented by a linear transformation $\mathcal{F}_{z(t)}$ after concatenation, so the overall modeling is formalized as: $ \\ \boldsymbol{z_l} = \underbrace{\mathcal{F}_{z(n)}\left(...\mathcal{F}_{z(1)}(\mathcal{F}_{z(0)}(\boldsymbol{z_g}\oplus \boldsymbol{e_{l_i}})\oplus \boldsymbol{e_{l_i}})...\oplus \boldsymbol{e_{l_i}}\right)}_{\text{totally } n \textit{ Broadcasting Layers}}$,
\end{enumerate}
Note that, to extend the broadcasting layer from one to multiple layers, we simply repeat it to broadcast the label signal to every layer of the label infuser as shown in Figure \ref{figlabel}. Hence we call this label infuser the \textit{Broadcasting Net}.

\subsection{Controllable Text Generation}
Finally, we feed the sampled local latent vector $\boldsymbol{z_l}$ to the decoder for word decoding. During training, the conditional modeling process of the decoder with labeled document $\boldsymbol{Y}$ can be formulized as:
\begin{equation}
    \begin{aligned}
    p(\boldsymbol{Y}\mid \boldsymbol{z_l}) &= p(y_1\mid \boldsymbol{z_l})\prod_{i = 2}^n p(y_i\mid y_{1:i-1}, \boldsymbol{z_l})
    \\&= p(y_1\mid \boldsymbol{z_l}) \prod_{i = 2}^n p(y_i\mid \boldsymbol{h_i}, \boldsymbol{z_l}),
    \end{aligned}
\end{equation}
where $\boldsymbol{h}_i$ is the $i$-th hidden state of the decoder that satisfies $\boldsymbol{h}_i = \text{Decoder}(\boldsymbol{h}_{i-1}, y_{i-1}, \boldsymbol{z_l})$. Unlike other PnP models that completely ignore \textbf{BaseAE} during plug-in training, we argue that to ensure the high efficiency of blending two separate domains (i.e., $\boldsymbol{e_l}$ and $\boldsymbol{z_g}$) and generating contexts with high quality from them, the decoder of BaseAE ought to take part in the optimization process and be regarded as a fine-tuning component in PluinAE. As a result, the reconstruction loss of PluginAE is on word token level instead of continuous latent level like PPVAE. This setting only activate less than a half parameters of BaseAE, and expends very little time (compared with baselines) but achieves considerable performances (see Section \ref{trainingcost} for details.).

Once we trained the PluginAE, we only need to input the sampled global latent vector from its prior $\boldsymbol{z_g}\sim N(0, I)$ and a control label $l_i$ to the model for controlled generation as shown in the Figure \ref{figmainmodel}.
\newcommand{\mysize}{0.30\linewidth}
\begin{table*}[]
\centering
\begin{tabular}{cccccc}
\toprule[1.5pt]
\textbf{Dataset} & \textbf{\#Voc. Size} & \textbf{\#Training Docs} & \textbf{\#Validation Docs} & \textbf{\#Test Docs} & \textbf{\#Avg. Length} \\ \hline
\textbf{Yelp}    & 10,005         & 200,000                  & 10,000                     & 10,000               & 9.11                   \\
\textbf{Titles}  & 30,005         & 128,000                  & 16,000                     & 16,000               & 9.27                   \\
\textbf{Yahoo}   & 30,005         & 400,000                  & 3,000                      & 3,000                & 9.93                   \\ \bottomrule[1.5pt]
\end{tabular}
\caption{Statistics of the preprocessed datasets.}
\label{tab:statistics}
\end{table*}

\subsection{Training Loss of PluginAE}
For the plug-in training, we employ the labeled corpus $\boldsymbol{Y}$ as training data. Since the PluginAE inherits the training scheme of auto-encoders, its training loss consists of two parts, namely reconstruction loss and latent regularization term. To ensure the consistency of models' learning process, we employ adversarial auto-encoder (AAE) loss for RNN-based model and variational auto-encoder (VAE) loss for BART-based model. The reconstruction loss of both types of model is the same, which is the cross-entropy loss between generated token logits and training sentences. The main difference between them is that, AAE loss uses the adversarial distance between latent prior and posterior for regularization \cite{makhzani2015adversarial,mai2020plug}, while VAE loss employs KL divergence. For RNN-based PluginAE, to avoid potential representation vanish issue in $\boldsymbol{z_l}$ \cite{zhao2017infovae}, we take a mutual information $I(\boldsymbol{Y};\boldsymbol{z_l})$ into consideration follow infoVAE \cite{zhao2017infovae,rezaabad2020learning}. It can be further factored in two items related to the KL divergence as follow:
\begin{equation}
    \begin{aligned}
        &I(\boldsymbol{Y};\boldsymbol{z_l}) \\&=\int q(\boldsymbol{Y}, \boldsymbol{z_l}) \log \frac{q(\boldsymbol{Y}, \boldsymbol{z_l})}{q(\boldsymbol{Y})q(\boldsymbol{z_l})} d \boldsymbol{Y} d \boldsymbol{z_l} \\ 
        &=\int q(\boldsymbol{Y}, \boldsymbol{z_l}) \log \frac{q(\boldsymbol{z_l}\mid \boldsymbol{Y})}{q(\boldsymbol{z_l})} d \boldsymbol{Y} d \boldsymbol{z_l} \\ 
        &=\int q(\boldsymbol{Y}, \boldsymbol{z_l})\left[\log \frac{q(\boldsymbol{z_l}\mid \boldsymbol{Y})}{p(\boldsymbol{z_l})}-\log \frac{q( \boldsymbol{z_l})}{p(\boldsymbol{z_l})}\right] d \boldsymbol{Y} d \boldsymbol{z_l} \\ 
        &=\int q(\boldsymbol{z_l}, \boldsymbol{Y})\log \frac{q(\boldsymbol{z_l}\mid \boldsymbol{Y})}{p(\boldsymbol{z_l})} d \boldsymbol{Y} d \boldsymbol{z_l} - \int q(\boldsymbol{z_l})\log \frac{q( \boldsymbol{z_l})}{p(\boldsymbol{z_l})}  d \boldsymbol{z_l}\\
        &=\mathbb{D}_{\mathrm{KL}}\left[q(\boldsymbol{z_l} \mid \boldsymbol{Y}) \| p(\boldsymbol{z_l})\right]- \mathbb{D}_{\mathrm{KL}}\left[q(\boldsymbol{z_l})\| p(\boldsymbol{z_l})\right].
    \end{aligned}
\end{equation}
And we approximate the KL term $\mathbb{D}_{\mathrm{KL}}(q(\boldsymbol{z_l}\mid \boldsymbol{Y})\|p(\boldsymbol{z_l}))$ by adversarial distance for RNN-based PluginAE. Finally, the holistic objectives of both types of PluginAE are specified as follows:
\begin{equation}
\begin{aligned}
    \max p(\boldsymbol{Y}, l)&\geq \mathbb{E}_{q(\boldsymbol{z_l} \mid \boldsymbol{Y})}[\log p(\boldsymbol{Y} \mid \boldsymbol{z_l})] - \lambda_{z_l} \mathcal{L}_{\boldsymbol{z_l}} \\ \text{where} \\
    \mathcal{L}_{\boldsymbol{z_l}}&= \left\{\begin{aligned}
    & Dist(\boldsymbol{z_l}, E(\boldsymbol{Y})) \\&-\lambda_{\text{info}} \mathbb{D}_{\mathrm{KL}}(q(\boldsymbol{z_l})\|p(\boldsymbol{z_l}))\quad \text{RNN-based}\\
    \\
    &\mathbb{D}_{\mathrm{KL}}(q(\boldsymbol{z_l}\mid \boldsymbol{{Y}})\|p(\boldsymbol{z_l}))\quad \text{BART-based}
    \end{aligned}\right.
\end{aligned}
\label{eqbaseloss}
\end{equation}
Here, the distance function in RNN-based loss is implemented as $Dist(\boldsymbol{z_l}, E(\boldsymbol{Y}))=\mathbb{E}_{p(\boldsymbol{Y}, l)}[-\log (1-D(E(\boldsymbol{Y})))]\\+\mathbb{E}_{p(\boldsymbol{z_g})}[-\log D(\boldsymbol{z_l})]$ with $D, E$ to be the discriminator and encoder respectively. In practice, for RNN-based PCAE, we use multi-layer neural network (as the discriminator) to calculate its latent regularization term by classifying random noise and latent vectors. For BART-based PCAE, we use reparamerization trick \cite{bowman2016generating} to parameterize the mean and log variance of the latent space (i.e., Gaussian) to calculate the KL divergence.
\section{Experimental Results and Analysis}
\begin{figure*}[ht]
\centering
\includegraphics[width=1.0\linewidth]{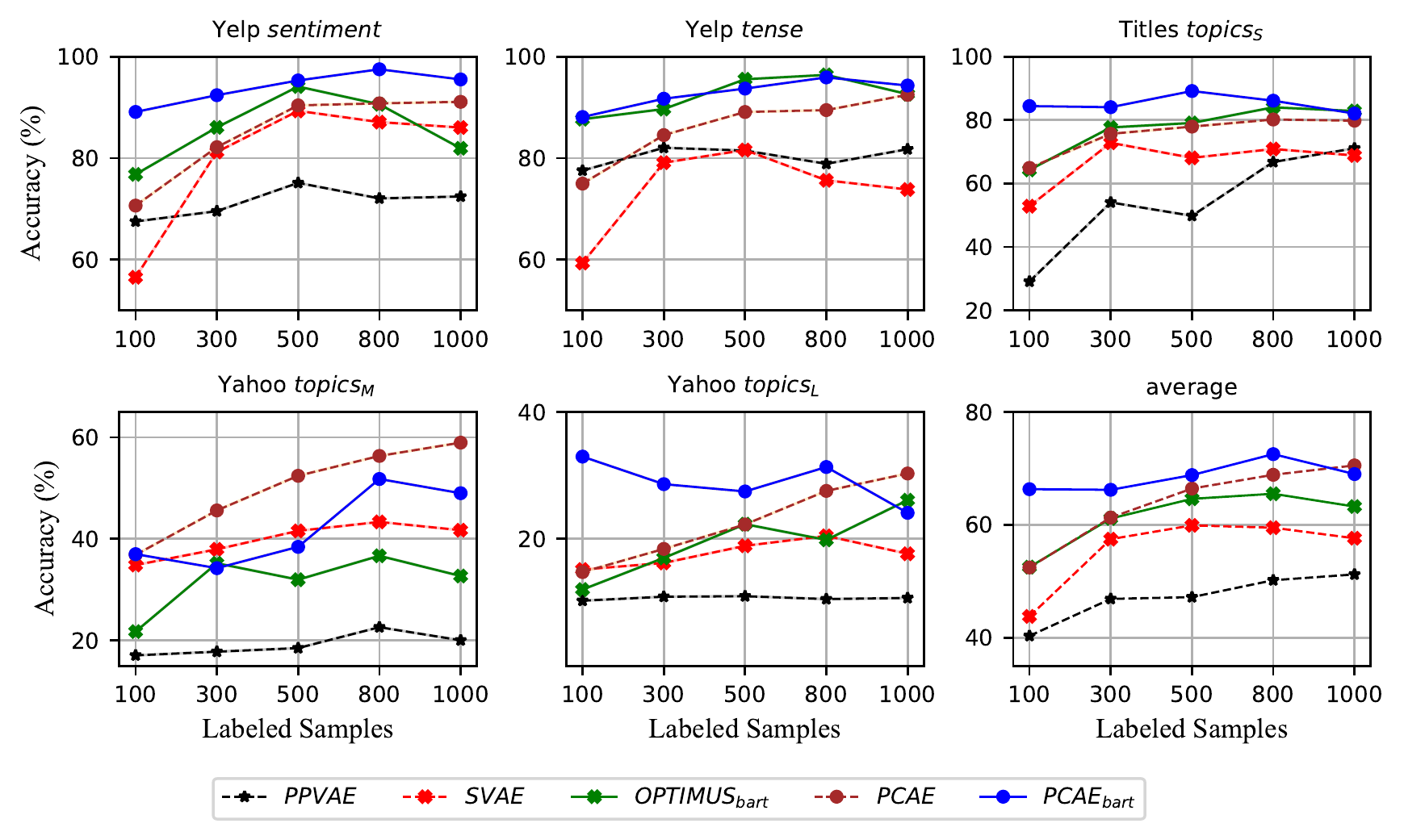}
\setlength{\abovecaptionskip}{-0.3cm}
\caption{Accuracy on five different tasks and averaged accuracy of them with varied number of labeled samples for each class. PCAE represents our framework under RNN framework with 12 layers in label fusion function $\Phi$ (\textit{Broadcasting Net}). $\text{PCAE}_{\text{bart}}$ represents our framework under pre-trained BART with 10 layers in label fusion function. We implemented OPTIMUS \cite{li2020optimus} under the same pre-trained BART framework as ours (denote as $\text{OPTIMUS}_{\text{bart}}$).}
\label{fig:acc}
\end{figure*}

\begin{table*}
\renewcommand\arraystretch{1.1}
\centering
\setlength{\tabcolsep}{2.5mm}{
\begin{tabular}{c|c|cccccc} 
\toprule[1.5pt]
                                        \textbf{Label Num.}             & \textbf{Models}         & \textbf{sentiment} & \textbf{tense} & $\textbf{topics}_{\textbf{S}}$ & $\textbf{topics}_{\textbf{M}}$                           & $\textbf{topics}_{\textbf{L}}$ & \textbf{Avg.}   \\ 
\hline\hline
 \multirow{3}{*}{\textbf{100 }}  & \textbf{SVAE}           & 56.53              & 59.34          & 52.86                          & 34.85                                                    & 15.18                 & 43.75           \\
                                                                        & \textbf{PPVAE}          & 67.53              & 77.56 & 29.10                          & 17.06                                                    & 10.26                          & 40.30           \\
                                                                        & $\textbf{OPTIMUS}_{bart}$        & 76.80              & 87.67          & 64.29                          & 21.73                                                    & 12.01                          & 52.44           \\ \cline{2-8}
                                                                        & \textbf{PCAE}           & 70.65     & 74.98          & 64.92                 & 36.86                                           & 14.77                          & 52.44  \\ 
                                                                        & $\textbf{PCAE}_{bart}$  & \textbf{89.10}     & \textbf{88.10}          & \textbf{84.40}                          & \textbf{37.00}                                           & \textbf{32.97}                          & \textbf{66.31}           \\ 
\cline{1-8}
                                        \multirow{3}{*}{\textbf{300 }}  & \textbf{SVAE}           & 81.20              & 79.10          & 72.82                          & 37.95                                                    & 16.21                          & 57.46           \\
                                                                        & \textbf{PPVAE}          & 68.55              & 82.04          & 54.05                          & 17.78                                                    & 10.89                          & 46.86           \\
                                                                        & $\textbf{OPTIMUS}_{bart}$        & 86.05              & 89.67          & 77.67                          & 35.20                                                    & 17.00                          & 61.28           \\ \cline{2-8}
                                                                        & \textbf{PCAE}           & 82.16     & 84.54 & 75.67                 & \textbf{45.61}                                           & 18.42                 & 61.28  \\
                                                                        & $\textbf{PCAE}_{bart}$  & \textbf{92.40}              & \textbf{91.70}          & \textbf{84.05}                          & 34.23                                                    & \textbf{28.63}                 & \textbf{66.20}           \\ 
\cline{1-8}
                                        \multirow{3}{*}{\textbf{500}}   & \textbf{SVAE}           & 89.31              & 81.60          & 68.10                          & 41.58                                                    & 18.91                          & 59.90           \\
                                                                        & \textbf{PPVAE}          & 75.12              & 81.49          & 49.82                          & 18.49                                                    & 10.97                          & 47.18           \\
                                                                        & $\textbf{OPTIMUS}_{bart}$        & 94.10              & \textbf{95.53} & 79.08                          & 31.97                                                    & 22.32                          & 66.42           \\ \cline{2-8}
                                                                        & \textbf{PCAE}           & 90.39     & 89.09 & 77.94                 & \textbf{52.43}                                           & 22.28                 & 66.42  \\
                                                                        & $\textbf{PCAE}_{bart}$  & \textbf{95.30}     & 93.70          & \textbf{89.15}                          & \begin{tabular}[c]{@{}c@{}} 38.43\\\end{tabular} & \textbf{27.47}                          & \textbf{68.81}  \\ 
\cline{1-8}
                                        \multirow{3}{*}{\textbf{800}}   & \textbf{SVAE}           & 87.10              & 75.63          & 70.88                          & 43.34                                                    & 20.48                          & 59.48           \\
                                                                        & \textbf{PPVAE}          & 72.07              & 78.89          & 66.78                          & 22.61                                                    & 10.52                          & 50.17           \\
                                                                        & $\textbf{OPTIMUS}_{bart}$        & 90.60              & \textbf{96.40}          & 84.00                          & 36.67                                                    & 19.84                          & 68.87           \\ \cline{2-8}
                                                                        & \textbf{PCAE}           & 90.79     & 89.46 & 80.17                 & \textbf{56.34}                                           & 27.56                 & 68.87  \\ 
                                                                        & $\textbf{PCAE}_{bart}$  & \textbf{97.50}              & 95.90 & \textbf{86.10}                          & 51.80                                           & \textbf{31.33}                 & \textbf{72.53}  \\ 
\cline{1-8}
                                        \multirow{3}{*}{\textbf{1000}}  & \textbf{SVAE}           & 86.02              & 73.82          & 68.85                          & 41.71                                                    & 17.66                          & 57.61           \\
                                                                        & \textbf{PPVAE}          & 72.44              & 81.76          & 71.10                          & 20.03                                                    & 10.69                          & 51.21           \\ 
                                                                        & $\textbf{OPTIMUS}_{bart}$        & 81.93              & 92.60          & \textbf{82.85}                          & 32.69                                                    & 26.10                          & 63.23           \\ \cline{2-8}
                                                                        & \textbf{PCAE}           & 91.09     & 92.48 & 79.80                 & \textbf{58.96}                                           & \textbf{30.31}                 & \textbf{70.52}  \\
                                                                        & $\textbf{PCAE}_{bart}$  & \textbf{95.50}              & \textbf{94.30} & 82.05                          & 49.00                                           & 24.11                          & 68.99           \\

\bottomrule[1.5pt]
\end{tabular}}
\caption{Accuracy on five different tasks and averaged accuracy of them with varied number of labeled samples for each class. All model settings are the same as models in Figure \ref{fig:acc}. We use \textbf{boldface} to indicate the best value.}
\label{tab:fig:acc}
\end{table*}

\begin{table*}
\renewcommand\arraystretch{1.1}
\centering
\setlength{\tabcolsep}{2.0mm}{
\begin{tabular}{c|c|cc|cc|cc|cc|cc} 
\toprule[1.5pt]
                                        \multirow{2}{*}{\textbf{Label Num.}} & \multirow{2}{*}{\textbf{Models}} & \multicolumn{2}{c|}{\textbf{sentiment}} & \multicolumn{2}{c|}{\textbf{tense }} & \multicolumn{2}{c|}{$\textbf{topics}_{\textbf{S}}$} & \multicolumn{2}{c|}{$\textbf{topics}_{\textbf{M}}$} & \multicolumn{2}{c}{$\textbf{topics}_{\textbf{L}}$}  \\ 
\cline{3-12}
                                                                            &                                  & \textbf{D-1 }  & \textbf{D-2 }          & \textbf{D-1 }  & \textbf{D-2 }       & \textbf{D-1 } & \textbf{D-2 }                       & \textbf{D-1 } & \textbf{D-2 }                       & \textbf{D-1 } & \textbf{D-2 }                       \\ 
\hline\hline
 \multirow{7}{*}{\textbf{100 }}       & \textbf{SVAE}                    & 1.38           & 20.68                  & 2.01           & 26.54               & 1.13          & 23.33                               & 0.29 & \textbf{9.29}                       & 0.12          & 4.87                                \\
                                       &                                       \textbf{PPVAE}                   & 2.91           & 27.60                  & 3.93  & \textbf{35.99}      & 1.59          & 17.49                               & 0.19          & 4.07                                & 0.18          & 3.43                                \\
                                       &    $\textbf{OPTIMUS}_{bart}$                 & 7.39           & 17.62                  & 6.11           & 12.27               & 4.28          & 10.90                               & 0.12          & 0.31                                & 0.45          & 2.73                                \\
\cline{2-12}
                                       &                                       $\textbf{PCAE}_{10}$               & 3.14           & 31.19                  & 3.52           & 33.98               & 2.49          & 23.46                               & 0.27          & 6.22                                & 0.40 & 4.99                       \\
                                       &                                       $\textbf{PCAE}_{15}$               & 2.92           & \textbf{31.38}         & 3.47           & 33.63               & 2.81 & \textbf{25.11}                      & 0.28          & 6.22                                & 0.39          & 4.96                                \\
                                       &                                       $\textbf{PCAE}_{bart10}$           & 9.98           & 25.54                  & \textbf{11.06}          & 28.34               & \textbf{5.43} & 11.53                               & \textbf{0.30} & 1.46                       & \textbf{1.28} & \textbf{7.45}                       \\
                                       &                                       $\textbf{PCAE}_{bart15}$           & \textbf{10.41}          & 27.17                  & 9.63           & 25.69               & 4.42          & 7.94                                & 0.17          & 1.05                                & 1.13          & 5.29                                \\ 
\cline{1-12}
                                        \multirow{7}{*}{\textbf{300 }}    &    \textbf{SVAE }                   & 1.23           & 22.86                  & 1.64           & 31.36               & 0.93          & 20.47                               & 0.20          & \textbf{7.02}                       & 0.11          & 4.82                                \\
                                       &                                       \textbf{PPVAE }                  & 2.50           & 28.75                  & 3.75 & 35.42               & 1.34          & 16.71                               & 0.19          & 3.91                                & 0.17          & 3.40                                \\
                                       &     $\textbf{OPTIMUS}_{bart}$        & 9.39           & 24.07                  & 11.10          & 24.15               & \textbf{5.65} & 14.06                               & 0.13          & 0.49                                & 0.60          & 3.22                                \\
\cline{2-12}
                                       &                                       $\textbf{PCAE}_{10}$               & 2.31           & 28.18                  & 2.95           & 35.23               & 1.96          & 23.70                               & 0.21          & 6.50                                & 0.33          & 5.26                       \\
                                       &                                       $\textbf{PCAE}_{15}$               & 2.58  & 28.93         & 2.99           & \textbf{37.06 }     & 1.93          & \textbf{23.89}                               & 0.21 & 6.47                                & 0.33 & 5.20                                \\
                                       &                                       $\textbf{PCAE}_{bart10}$           & \textbf{14.08} & \textbf{40.31}         & \textbf{12.74} & 22.53               & 5.59          & 16.73                      & \textbf{0.36} & 1.82                       & 1.47          & 7.85                                \\
                                       &                                       $\textbf{PCAE}_{bart15}$           & 12.86          & 35.94                  & 12.08          & 19.76               & 4.80          & 11.07                               & 0.17          & 0.90                                & \textbf{4.14} & \textbf{9.98}                       \\ 
\cline{1-12}
                                        \multirow{7}{*}{\textbf{500 }}     &   \textbf{SVAE }                   & 1.34           & 25.23                  & 1.83           & 34.55               & 0.78          & 19.19                               & 0.16          & 5.89                                & 0.11          & 4.73                                \\
                                       &                                       \textbf{PPVAE }                  & 3.53 & 34.83                  & 3.73 & 35.23               & 1.13          & 15.90                               & 0.18          & 3.76                                & 0.17          & 3.37                                \\
                                       &      $\textbf{OPTIMUS}_{bart}$        & 10.96          & 26.37                  & 11.84          & 24.74               & 6.52          & 15.58                               & 0.08          & 0.29                                & 0.89          & 4.11                                \\
\cline{2-12}
                                       &                                       $\textbf{PCAE}_{10}$               & 2.19           & 28.65                  & 2.69           & 31.60               & 2.15          & 24.52                               & 0.23          & 6.43                                & 0.38 & 5.03                                \\
                                       &                                       $\textbf{PCAE}_{15}$               & 2.30           & 34.86         & 2.66           & \textbf{36.32 }     & 2.10          & \textbf{25.28}                               & 0.23 & \textbf{6.59}                       & 0.37          & 5.05                       \\
                                       &                                       $\textbf{PCAE}_{bart10}$           & 14.21          & 40.91                  & \textbf{17.72} & 33.30      & 8.65          & 19.53                      & 0.38          & 1.87                                & \textbf{1.60} & \textbf{7.67}                       \\
                                       &                                       $\textbf{PCAE}_{bart15}$           & \textbf{15.41} & \textbf{42.58}         & 7.16           & 11.14               & \textbf{8.98} & 17.51                               & \textbf{0.39} & 1.92                       & 0.64          & 3.55                                \\ 
\cline{1-12}
                                        \multirow{7}{*}{\textbf{800 }}   &     \textbf{SVAE}                    & 1.75           & 25.42                  & 1.95           & 35.32               & 0.71          & 19.35                               & 0.15          & 5.32                                & 0.10          & 4.69                                \\
                                       &                                       \textbf{PPVAE}                   & 2.32  & 31.71                  & 3.57  & 34.83               & 0.99          & 15.26                               & 0.18          & 3.62                                & 0.17          & 3.34                                \\
                                       &     $\textbf{OPTIMUS}_{bart}$        & 12.39          & 31.06                  & 14.34          & 32.25               & 7.13          & 17.43                               & 0.10          & 0.31                                & 0.43          & 2.64                                \\
\cline{2-12}
                                       &                                       $\textbf{PCAE}_{10}$               & 2.12           & 29.71                  & 2.96           & 32.62               & 2.52         & 24.27                               & 0.26          & 6.39                                & 0.44 & 5.45                                \\
                                       &                                       $\textbf{PCAE}_{15}$               & 2.25           & 34.05         & 2.80           & \textbf{37.74}      & 2.45          & \textbf{24.90}                      & 0.27 & \textbf{6.45}                       & 0.44          & 5.48                       \\
                                       &                                       $\textbf{PCAE}_{bart10}$           & \textbf{15.64} & 46.79                  & \textbf{15.51} & 28.60               & \textbf{9.95} & 22.25                      & \textbf{0.40} & 2.14                       & \textbf{1.66} & \textbf{8.14}                       \\
                                       &                                       $\textbf{PCAE}_{bart15}$           & 15.38          & \textbf{48.19}         & 15.34          & 29.41               & 6.99          & 14.90                               & 0.12          & 0.75                                & 0.81          & 4.31                                \\

\bottomrule[1.5pt]
\end{tabular}}
\caption{Distinct-1 and Distinct-2 (refer to D-1 and D-2 respectively) of different models with varied number of labeled samples for each class. Real values are the presented ones divided by 100. $\textbf{PCAE}_{n}$ and $\textbf{PCAE}_{bartn}$ means the proposed RNN-based PCAE and BART-based PCAE model with $n$ broadcasting layers respectively. We use \textbf{boldface} to indicate the best value.}
\label{tab_dis}
\end{table*}

\subsection{Implementation Details}
\subsubsection{RNN-based Implementation Details}
For BaseAE, we formally followed the settings in  \cite{shen2020educating}. For three datasets to be trained, we mainly focused the short text generation and set the maximum vocabulary size to 10,000 for Yelp, 30,000 for Yahoo dataset and Titles dataset. Statistics for pre-training corpus are listed in Table \ref{tab:statistics}. Word embedding dimension was 512. The encoder and decoder of BaseAE were bi-directional LSTM \cite{hochreiter1997long} and plain LSTM severally,
and both with a hidden size of 1024 per direction. The size of global latent code was 128. The dimension of hidden state in the discriminator for latent distance measurement was set to 512. The noise function was taken from  \cite{shen2020educating}, we set word drop rate to 0.3. The weight of discriminator loss $\lambda_{\text{adv}}$ was set to 10. For optimization, we utilized Adam \cite{kingma2014adam} with learning rate $5\times 10^{-4}$ and a batch size of 256. We trained 50 epochs for three datasets, and stored model parameter weights on the epoch performs the best on validation set.

For PluginAE, the label embedding size $n$ was 8. The label \textit{Broadcasting Net} was implemented with pure linear functions and concatenations. During training, we activated the decoder, look-up linear function from decoder to word probability in the decoding section. And we also activated the linear function from latent codes to decoder embedding.
As for optimizer, we selected Adam with learning rate $1\times 10^{-4}$ and a batch of 80 samples as input. We trained our model on each task until it converges. Through cross-validations, we chose the weight of adversarial loss $\lambda_{\text{adv}}$ and latent regulator $\lambda_{\text{info}}$ to be 30 and 50 respectively. For text generation, we chose categorical sampling with decoding temperature to be 0.8 according to ablation experiments.

As for the formal implementation of $\mathbb{E}_{p(\boldsymbol{Y}, l)}[\mathbb{D}_{\mathrm{KL}}(q(\boldsymbol{z_l} \mid \boldsymbol{Y}) \| p(\boldsymbol{z_l}))]$ approximation in the PluginAE training loss. We utilized another divergence Maximum Mean Discrepancy (MMD) \cite{gretton2006kernel,li2015generative} to efficiently optimize  $\mathbb{D}_{\mathrm{KL}}(q(\boldsymbol{z_l}) \| p(\boldsymbol{z_l}))$ term. MMD is widely used for quantifying the distance between two distributions using the kernel trick. The MMD between two distributions $q$ and $p$ is:
\begin{equation}
    \begin{aligned}
        \mathbb{D}_{\mathrm{MMD}}(q \| p) &=\mathbb{E}_{p(z), p\left(z^{\prime}\right)}\left[k\left(z, z^{\prime}\right)\right]\\&-2 \mathbb{E}_{q(z), p\left(z^{\prime}\right)}\left[k\left(z, z^{\prime}\right)\right] \\ &+\mathbb{E}_{q(z), q\left(z^{\prime}\right)}\left[k\left(z, z^{\prime}\right)\right],
    \end{aligned}
\end{equation}
the function $k(z, z^{\prime})$ here is a definite kernel, and we chose it to be Gaussian. 
\subsubsection{BART-based Implementation Details}
We utilized BART encoder and decoder as encoder and decoder of our AE model respectively. For BaseAE structure, we used the pre-trained tokenizer with the vocabulary size of 50,265. To perform BART under the paradigm of auto-encoder, a latent space is required to connect encoder and decoder during training. We derive this latent space using the mean pooling of the output of encoder and feed it to decoder as cross attention input, which is similar to OPTIMUS \cite{li2020optimus}. We loaded the medium-sized pre-trained weight of BART \footnote{\url{https://huggingface.co/facebook/bart-base}} which consists of 6 transformer layers for encoder and decoder separately. The size of latent code was set to 128 like RNN-based ones. As for BaseAE training, we employed AdamW \cite{loshchilov2018decoupled} optimizer with learning rate $1\times 10^{-4}$ for three corpus. During training, we followed OPTIMUS to utilized free KL threshold \cite{li2019surprisingly,pelsmaeker2020effective}, the threshold was set to 0.1 and our model was trained with 4 cycles of KL annealing from 0 to 1 using cyclic KL annealing technique \cite{fu2019cyclical}. We finetuned the BART VAE with batch size 64 for 8, 10, 10 epochs for Yelp, Titles and Yahoo datasets respectively, which took around 2, 4 and 3 hours to train.

For PluginAE structure, we only added label infuser based on BaseAE model. The structure of label infuser is exactly the same as RNN-based PCAE models. For PluginAE training, we trained the model with the VAE training objectives and set the free KL threshold to 0.1 for training consistency. During training, we activated the decoder and label Broadcasting components. As for optimizer, we selected AdamW with a batch of 32 samples as input. Moreover, since PLMs are sensitive to learning rate during training, we chose different learning rate for different tasks according to their classification performance, i.e., $1\times 10^{-4}$ for tense, $\text{topic}_{L}$ tasks, $3\times 10^{-4}$ for sentiment, $\text{topic}_{S}$ and $\text{topic}_{M}$ tasks.  We trained our model on each task until it achieves the highest controllability (i.e., the highest accuracy on each task). For text generation, we chose top-$k$ nucleus sampling strategy \cite{holtzman2019curious} for decoding with $k=50$ and $p=1.0$ and sampling temperature to be $1.0$. We generated 500 sentences for each class in every task for further evaluations.
\subsubsection{Classifier Implementation Details}
For the classifier we applied for text attribute classification, we employed a bi-directional LSTM with one layer and a hidden size of $256$. The word embedding size was set to 128. As for optimization, we employed SGD \cite{bottou2010large} with learning rate 0.01. We trained the classifier on five tasks with 5,000 labeled samples for each class and 50 for a batch until the loss converges. We saved the parameter weights of model performs the best on validation set during training.

\subsection{Datasets}
We conducted controlled experiments on different tasks to quantify the benefits of the various aspects of our approach. Specifically speaking, we followed previous works and carried out related tasks on three datasets: Yelp review \cite{shen2017style}, Titles \cite{fu2018style} and Yahoo Question \cite{yang2017improved}. We chose five tasks from these three datasets, all with text semantic labels. Their detailed class descriptions are presented below:
\begin{itemize}
    \item Yelp sentiment: $2$ classes. Positive, Negative.
    \item Yelp tense: $2$ classes. Present, Past. We make the same partition as described in  \cite{shen2020educating}.
    \item Titles $\text{topics}_{\text{S}}$: $4$ classes. Business, Science \& Technology, Entertainment, Health.
    \item Yahoo $\text{topics}_{\text{M}}$: $6$ classes. Society \& Culture, Science \& Mathematics, Health, Education \& Reference, Computers \& Internet, Sports.
    \item Yahoo $\text{topics}_{\text{L}}$: $10$ classes. $6$ from Yahoo $\text{topics}_{\text{M}}$ task with the addition to Business \& Finance, Entertainment \& Music, Family \& Relationships, Politics \& Government.
\end{itemize}
We respectively sampled $100, 300, 500, 800, 1000$ examples from each classes for every task as plug-in training sets. The detailed statistical summary of three datasets for BaseAE pre-training is reported in Table \ref{tab:statistics}.

\subsection{Baselines}
For RNN-based implementation of our model, the first comparison focuses on a model that, likes ours, is plug-and-play. PPVAE \cite{duan2020pre} generates controlled texts by feeding sentences with a heavy bias i.e., all from the same class. For multi-class controlled generation, PPVAE also takes negative samples (those not in the current desired category) and produce a negative loss to enhance its overall ability. That is to say, for $n$ conditions (classes), PPVAE needs to produce $n$ plug-in VAEs to be controlled, besides, at every training procedure it normally requires to encode all the given samples so that negative loss can be computed. Another baseline model SVAE \cite{kingma2014semi} follows the end-to-end semi-supervised framework. It incorporates a classifier to provide conditional distribution for unlabeled data. Note that, to make it equal, we implement all baselines \textbf{under the same BaseAE}, and update the decoder parameters in PPVAE to make the competition impartial. Since other models like CTRL-GEN \cite{hu2017toward} has been proven much inferior in similar tasks \cite{duan2020pre}, we did not take them for comparison.

For BART-based implementation of our model, we took OPTIMUS \cite{li2020optimus} as the baseline model. OPTIMUS is a strong baseline that reached unparallel language modeling ability including controllability, its controllable training requires a two-stage training process, and we used the same BaseAE of ours as its first stage tuning model out of fairness. OPTIMUS introduced a discriminator on latent space for adversarial training (to distinguish the real latent variable and Gaussian noise). We employed the same discriminator structure as in OPTIMUS with latent space dimension of 128. We denote BART-based OPTIMUS as $\text{OPTIMUS}_{bart}$.


\begin{figure*}[ht]
\centering
\subfigure[Yelp sentiment]{
\includegraphics[width=\mysize]{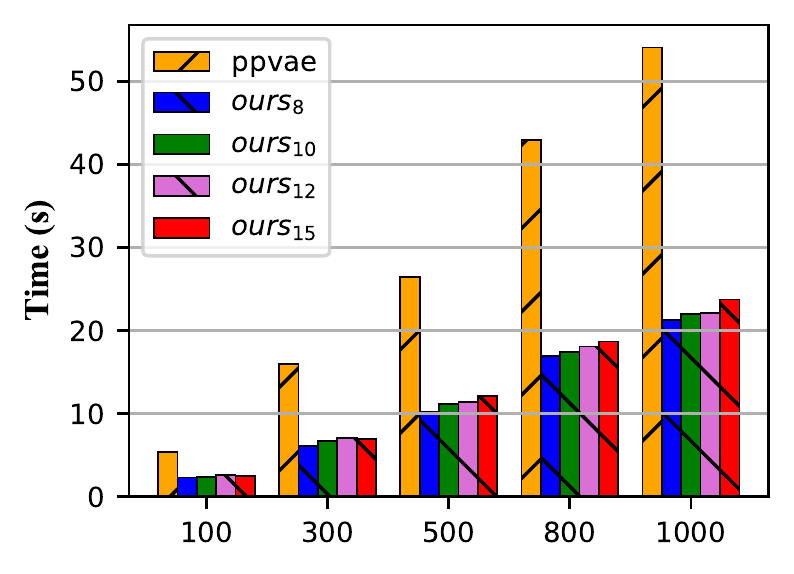}
}
\subfigure[Titles $\text{topics}_{\text{S}}$]{
\includegraphics[width=\mysize]{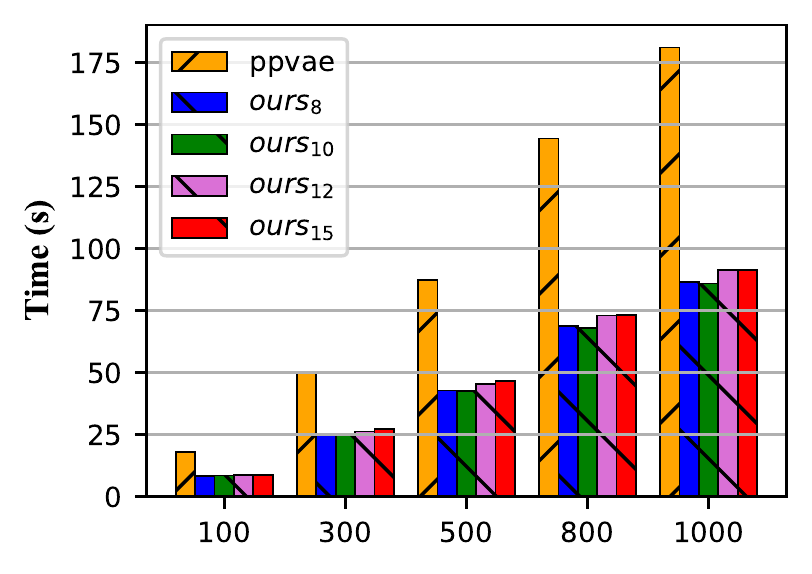}
}
\subfigure[Yahoo $\text{topics}_{\text{L}}$]{
\includegraphics[width=\mysize]{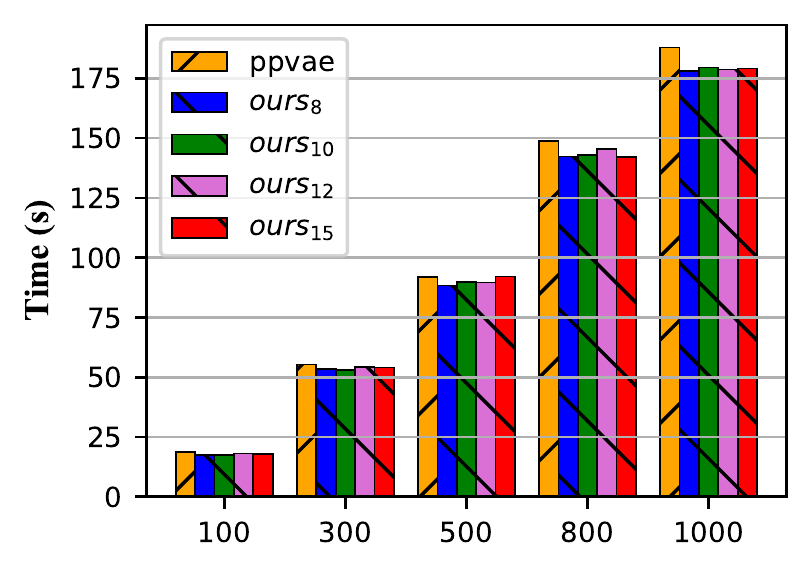}
}
\setlength{\abovecaptionskip}{0.1cm}
\caption{Training time (Y axis, counted in second) for plug-in modules with varied number of labeled samples (X axis) for each class on three tasks from three different datasets.}
\label{fig:training_time}
\end{figure*}
\begin{table}[]
\centering
\begin{tabular}{c|cc}
\toprule[1.5pt]
\textbf{Models} & \multicolumn{2}{c}{\textbf{Time Cost}} \\ \hline
\textbf{SVAE}   & \multicolumn{2}{c}{$\sim$1.1h (every time)}          \\
\textbf{PPVAE}  & 1h (only once) & 55.71s (Plug-in) \\
\textbf{Ours}   & 1h (only once) & \textbf{37.18s} (Plug-in) \\
\bottomrule[1.5pt]
\end{tabular}
\caption{Averaged time cost for training with $300$ labeled data for each class on RNN-based systems.}
\label{tab_avgtime}
\end{table}
\subsection{Evaluations and Analysis}
\subsubsection{Controllability} \label{sec:control}
To evaluate which degree of controlment the proposed model can obtain, we conducted experiments of text attribute classification. The classifier was trained with $5,000$ samples for each category. From the results in Figure \ref{fig:acc} and Table \ref{fig:acc}, we could draw the following conclusions: (1) Our RNN-based model is evidently superior to baseline PPVAE that follows the PnP paradigm in all circumstances. Also, in most situations, our model outperforms SVAE, which is an end-to-end model that needs a full-training every time. The last figure of averaged accuracy well confirms that PCAE is a better performer in controllable generation than baselines. (2) In some cases, the proposed model is capable of reaching comparatively high accuracy even with few available labels with either RNN or BART (e.g., RNN-based model reaches over $80\%$ accuracy with only $300$ labeled samples for each class on Yelp tense and sentiment). (3) When model's encoder and decoder are not very strong (i.e., RNN trained from the scratch), in tasks with too many categories (i.e., no less than $4$ classes), models with explicit label signals (i.e., our model and SVAE) outperforms PPVAE distinctly and the proposed RNN-based model outperforms SVAE in most circumstances. We attribute these phenomenons to an effective label fusion network of ours compared with baselines. (4) BART-based models achieve generally higher accuracy than RNN-based models (except in $\text{topics}_{\text{M}}$ task), which can be ascribed to applying powerful PLM encoder\&decoder. (5) From the averaged result, among all baselines, the proposed $\text{PCAE}_{bart}$ reaches the best performance in the most cases and comparable accuracy results in the rest situations.
\begin{figure*}[ht]
\centering
\subfigure[]{
\includegraphics[width=1.0\linewidth]{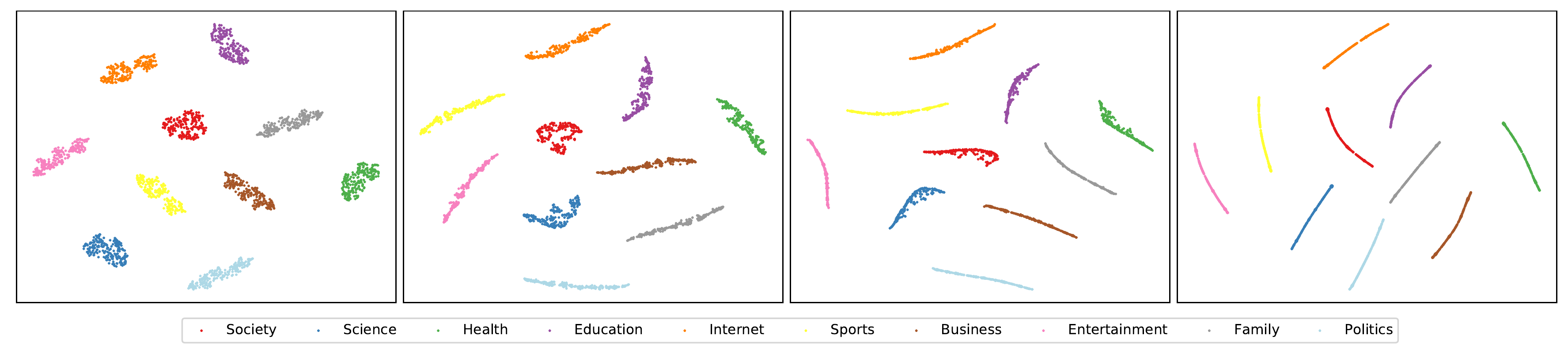}}
\subfigure[]{
\includegraphics[width=1.0\linewidth]{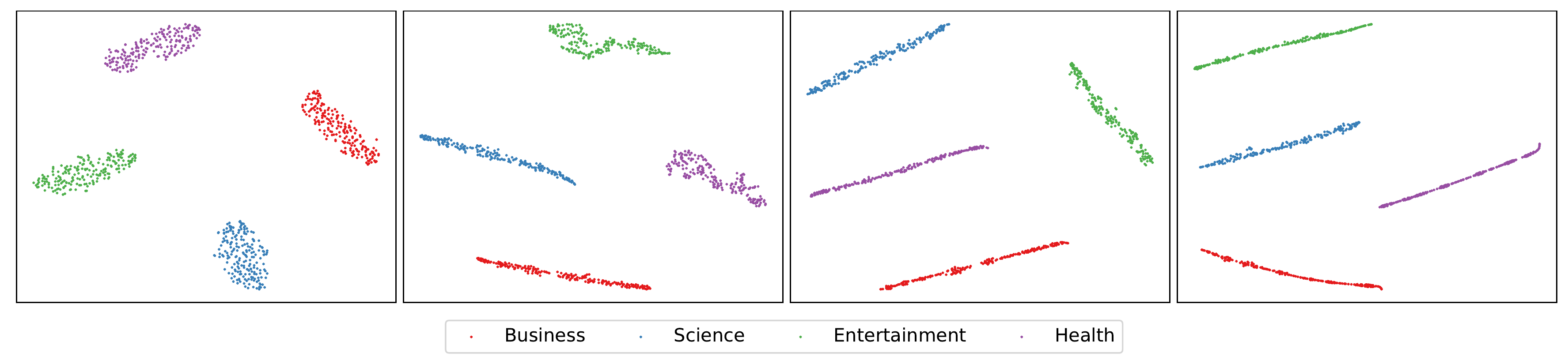}}
\setlength{\abovecaptionskip}{-0.11cm}
\caption{Visualization of the designed prior of local latent $\boldsymbol{z_l}$ from RNN-based PCAE with varied layers in \textit{Broadcasting Net} on (a) Yahoo $\text{topics}_{\text{L}}$ task and (b) $\text{topics}_{\text{S}}$ task. The number of broadcasting layer is chosen from 5, 8, 10, 12 in succession.}
\label{fig:tsnevis}
\end{figure*}
\begin{table*}
\centering
\begin{tabular}{l|c|l|l} 
\toprule[1.5pt]
                                               & \textbf{Task}                                   & \textbf{Condition}              & \textbf{Generated Examples}                                                                                      \\ 
\hline\hline
\multirow{8}{*}{\rotcell{\textbf{RNN-based }}} & \multirow{2}{*}{\textbf{sentiment}}             & \textbf{Negative}               & $\bullet$this main experience was totally short and no good .                                                    \\
                                               &                                                 & \textbf{Positive}               & $\bullet$good experience in the area                                                                             \\ 
\cline{2-4}
                                               & \multirow{2}{*}{\textbf{tense}}                 & \textbf{Past}                   & $\bullet$the fries were interesting , but not overcooked by the toppings .                                       \\
                                               &                                                 & \textbf{Present}                & $\bullet$the place is always good and the brunch is the beginning .                                              \\ 
\cline{2-4}
                                               & \multirow{4}{*}{$\textbf{topics}_{\textbf{S}}$} & \textbf{Entertainment}          & $\bullet$kate gomez sparks justin bieber engagement in an ring                                                   \\
                                               &                                                 & \textbf{Business}               & $\bullet$linkedin to oracle will offer hiding on data today                                                      \\
                                               &                                                 & \textbf{Technology}             & $\bullet$lg launches mini spots of a now \$199 on the september                                                  \\
                                               &                                                 & \textbf{Health}                 & $\bullet$sierra of un: scientists fight ebola virus family                                                       \\ 
\hline\hline
\multirow{8}{*}{\rotcell{\textbf{BART-based}}} & \multirow{2}{*}{\textbf{sentiment }}            & \textbf{\textbf{Negative}}      & $\bullet$furthermore, the food quality does not meet the price.                                                  \\
                                               &                                                 & \textbf{\textbf{Positive}}      & $\bullet$really good burgers and the oak stout beer is really smooth and tasty.                                  \\ 
\cline{2-4}
                                               & \multirow{2}{*}{\textbf{tense }}                & \textbf{\textbf{Past}}          & $\bullet$the paella was really really really good and the manhattans did not disappoint.                         \\
                                               &                                                 & \textbf{\textbf{Present}}       & $\bullet$they compete with each other and \textbf{do} not care about you!                                        \\ 
\cline{2-4}
                                               & \multirow{4}{*}{$\textbf{topics}_{\textbf{S}}$} & \textbf{\textbf{Entertainment}} & $\bullet$is hollywood to blame? or journalism? a battle on twitter                                               \\
                                               &                                                 & \textbf{\textbf{Business}}      & $\bullet$tinder may not be worth \$5b, but it's way more valuable than you think                                 \\
                                               &                                                 & \textbf{\textbf{Technology}}    & $\bullet$vintage-look electric car that could replace horse-drawn carriages\textcolor[rgb]{0.502,0.502,0.502}{}  \\
                                               &                                                 & \textbf{\textbf{Health}}        & $\bullet$2780-calorie french toast: cheesecake factory tops the (calorie) charts                                 \\
\bottomrule[1.5pt]
\end{tabular}
\caption{Conditional examples generated by PCAE on three tasks for qualitative analysis.}
\label{tab:texts}
\end{table*}

\subsubsection{Diversity}\label{sec:diversity}
Intuitively, texts with high controllability often face with the conundrum of low diversity. AE-based works have been widely explored and revealed the capacity to generate diverse contents \cite{wang2017diverse,razavi2019generating}. We measured distinct $n$-grams (normalized by the length of text) as in \cite{li2016diversity}:
\begin{equation}
\centering
    \text{Distinct-}n = \frac{\text{unique }n\text{-grams}}{N},
\end{equation}
where $N$ is the number of generated words. Higher the Distinctive scores are, less likely the model produces ``dull texts''. We report the ratio of unique $1$-gram and $2$-gram values (refer to D-1 and D-2 respectively) of 5k sentences for each category from any model on five tasks in Table \ref{tab_dis}. (1) For both RNN-based and BART-based models, our model is able to generate more diverse sentences than baselines in most cases. Especially on $\text{topics}_{\text{M}}$ and $\text{topics}_{\text{L}}$, the gains in diversity from our model to other methods are significantly higher (PCAE is twice as better as PPVAE or SVAE on D-1, $\text{PCAE}_{bart}$ is almost triple as better as $\text{OPTIMUS}_{bart}$ on D-2). (2) Introducing pre-trained BART to our model can largely bring up the D-1 values. We ascribe it to the larger vocabulary size of pre-trained BART (i.e., 50,265 for BART by default and maximum 30,000 for RNNs by presetting). (3) BART-based PCAE has little or non improvement on D-2 compared with RNN-based models. This is because $\text{PCAE}_{bart}$ reaches a higher degree of controllability, which is naturally contradicts to text diversity. That is, as the increase of  \textit{broadcasting layer}, $\text{PCAE}_{bart}$ may produce more structured local latent space, which is in favor of generating controllable sentences but against high diversity in the latent representations. This can be explained as the $\text{PCAE}_{bart10}$ always generates more diverse texts than $\text{PCAE}_{bart15}$ in the table.
\subsubsection{Training Cost}\label{trainingcost}
Can our model be applicable in real scenarios? To answer that question, the training time of plug-in modules should be paid great attention to. We make comparison between RNN-based PCAE and PPVAE, which are plug-and-play. In detail, we recorded times that every model consumed until it converged, and all models were trained on the same machine with one TITAN X GPU successively. For every picture from Figure \ref{fig:training_time}, we draw the time consumption in seconds of RNN-based models. We have the following conclusions: (1) Our model outperforms PPVAE distinctively in all circumstances (less than a half of the training time compared with PPVAE in both model settings). This demonstrates the effectiveness of our model for being a more pragmatic tool for controllable text generation. (2) In cases of two tasks from Yahoo dataset, PPVAE is less disadvantageous in time cost than other tasks compared with ours. We argue that these cases should be analyzed combined with the accuracy metric: PPVAE can not handle situations where too many classes come in at one time, thus quits learning to be controlled (e.g., PPVAE holds steady accuracy of $20\%$ and $15\%$ for $\text{topics}_{\text{M}}$ and $\text{topics}_{\text{L}}$ in Figure \ref{fig:acc}) and converges without acquiring enough knowledge from the biased data. 
(3) Varying the number of broadcasting layers in the label infuser dose not influence the training time apparently. Because each broadcasting layer contains a small number of parameters to be updated. We also present the averaged time cost of baseline models compared with our model in Table \ref{tab_avgtime}. SVAE and $\text{OPTIMUS}_{bart}$ as end-to-end model and two-stage fine-tuning model respectively are much slower and unpractical to be controllable when coming across too many conditions.
\begin{figure*}[ht]
\centering
\subfigure[Yelp sentiment]{
\includegraphics[width=0.3\linewidth]{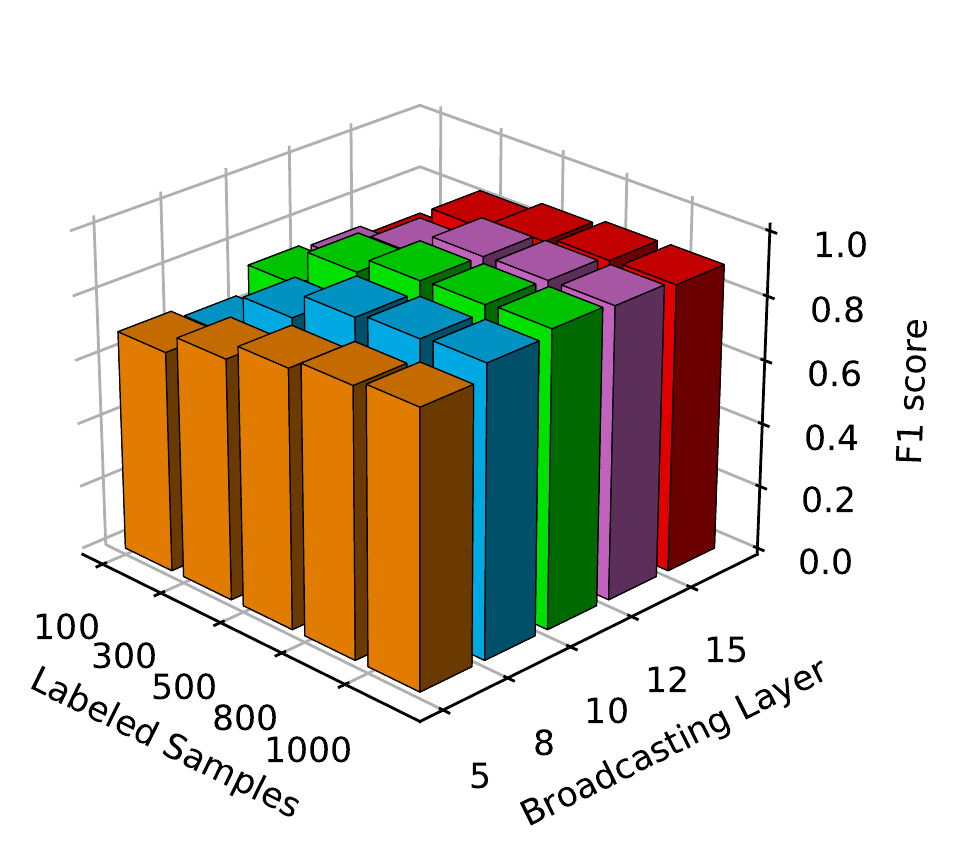}}
\subfigure[Yelp tense]{
\includegraphics[width=0.3\linewidth]{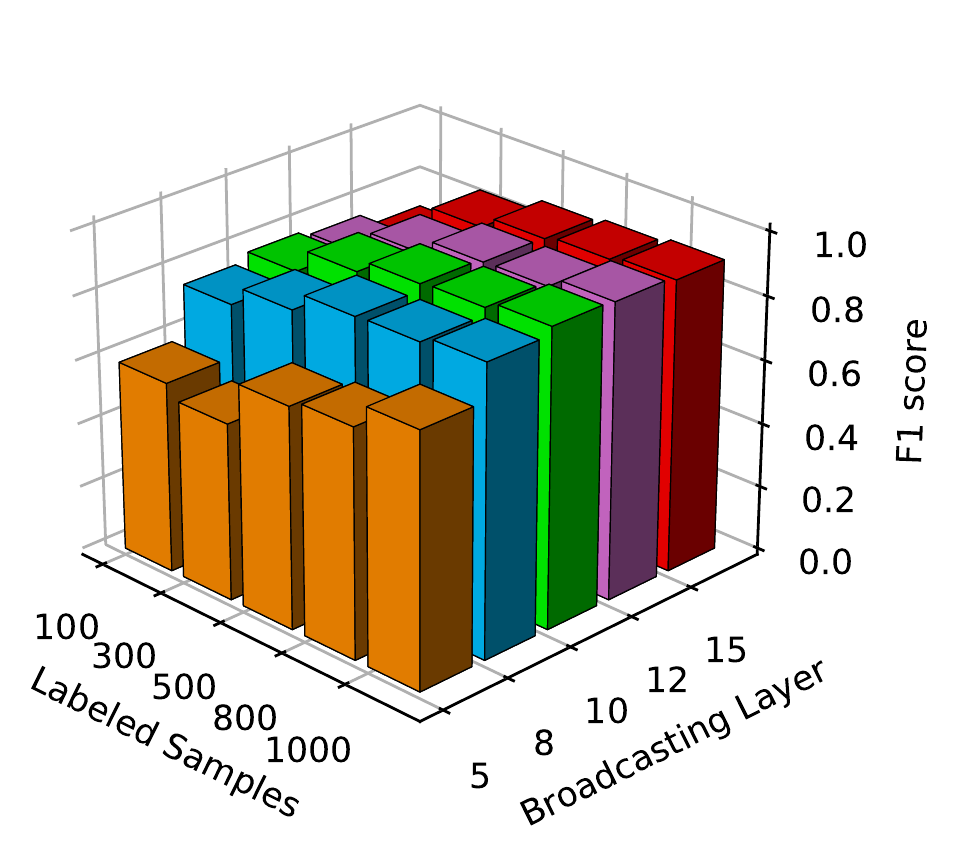}}
\subfigure[$\text{topics}_{\text{S}}$]{
\includegraphics[width=0.3\linewidth]{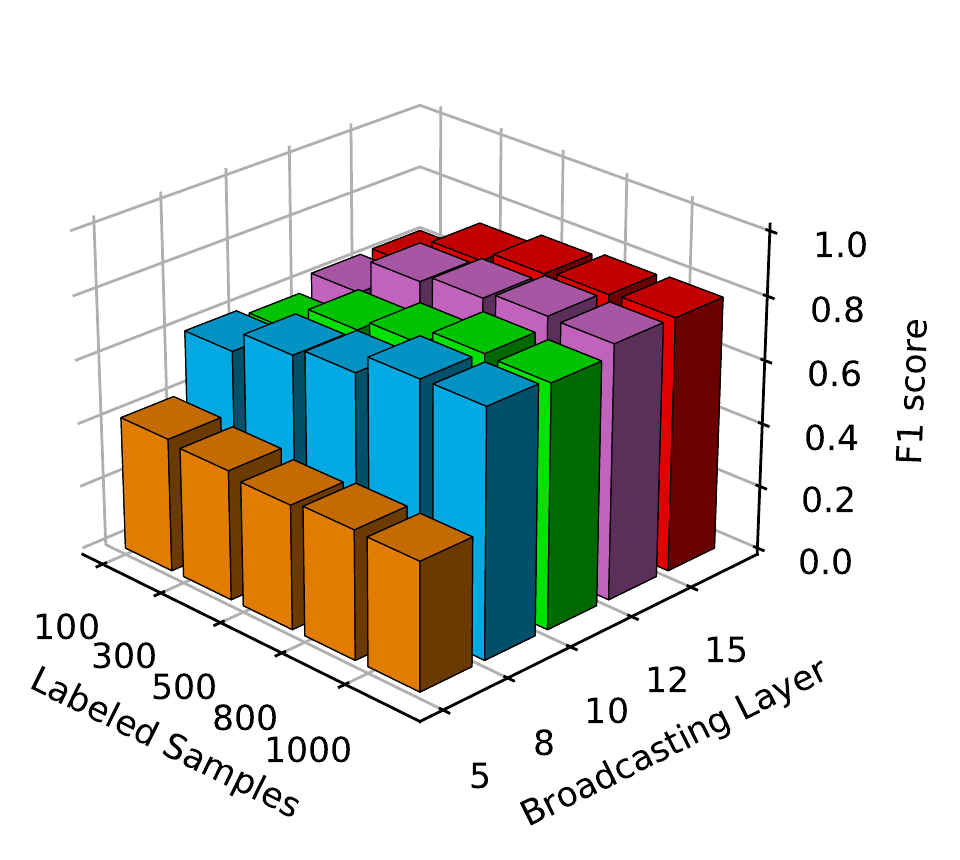}}
\subfigure[$\text{topics}_{\text{M}}$]{
\includegraphics[width=0.3\linewidth]{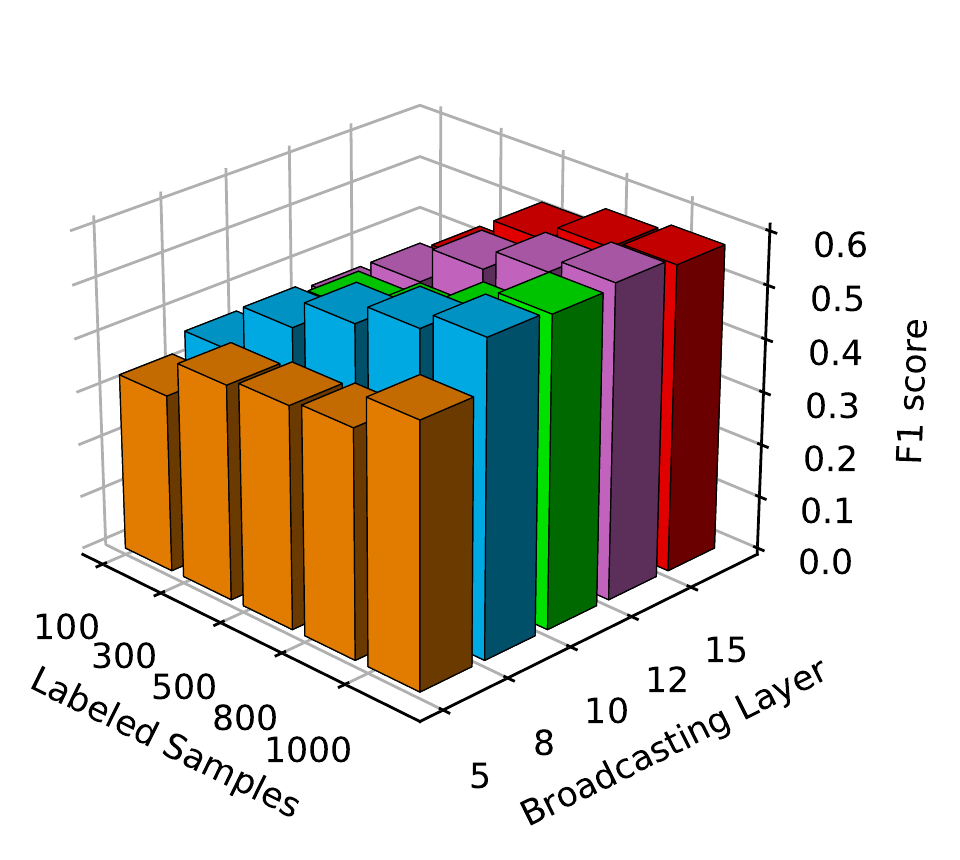}}
\subfigure[$\text{topics}_{\text{L}}$]{
\includegraphics[width=0.3\linewidth]{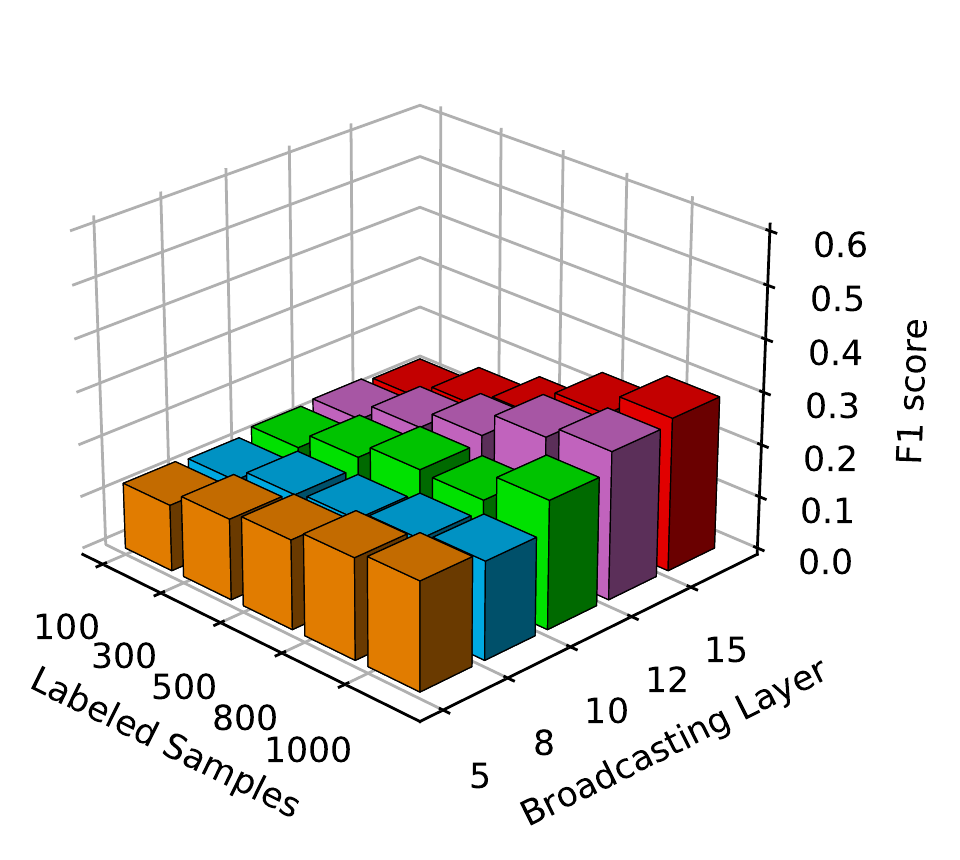}}
\subfigure[averaged]{
\includegraphics[width=0.3\linewidth]{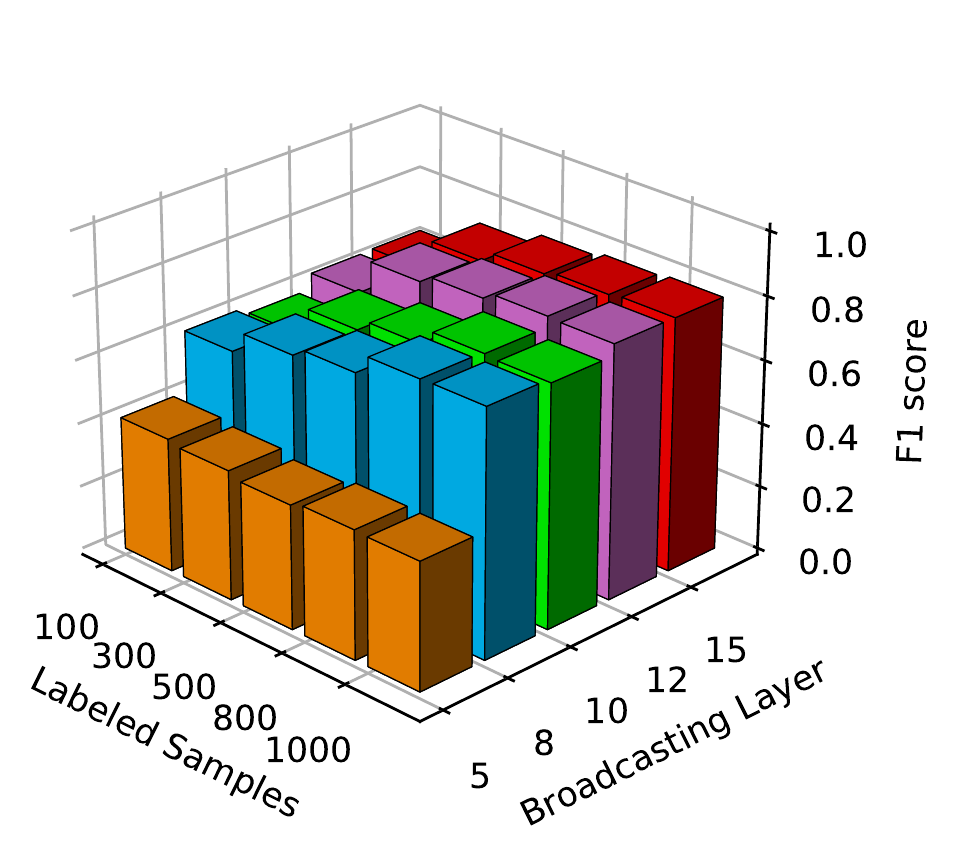}}
\setlength{\abovecaptionskip}{-0.11cm}
\caption{Macro-F1 score for generated controllable text classification with regard to varying broadcasting layer and labeled samples for each class based on RNN-based PCAE.}
\label{fig:ablationln}
\end{figure*}
\subsubsection{Justification for Latent Optimization}
To illustrate that the whole plug-in training process is more inclined to enhance the controlled expression in the latent field (i.e., helps $\boldsymbol{z_g}$ towards meaningful and structured $\boldsymbol{z_l}$), updating decoder is only an auxiliary tool for producing more fluent sequential content. We resort to visualize the input of model decoder using T-SNE \cite{van2008visualizing}, the visualized input corresponds to the prior of local latent code $\boldsymbol{z_l}$. As shown in Figure \ref{fig:tsnevis}, under the setting of RNN-based PCAE on $\text{topics}_{\text{L}}$ task and $\text{topics}_{\text{S}}$ task, the factitious local latent prior with given labels are well separated. Priors from \textit{Broadcasting Net} with added layers show more compact and structured clustering, which further verifies our claim in Sec. \ref{sec:diversity} for the discussions about generated text diversity. This phenomenon also justifies that the \textit{Broadcasting Net} as our label infuser function $\Phi$ efficiently elevates the capacity in the learnt hidden representations. From another view, PCAE produces controllable sentences by focusing on $\boldsymbol{z_g}$ to $\boldsymbol{z_l}$ rather than updating the parameters in decoder. We can also observe that, topics that are intuitively more correlated in real-life are actually closer in clustering (e.g., society is always in the center, family locates close to education).

\subsubsection{Generated Sentences}
We present texts with different conditions on three tasks (sentiment, tense, $\text{topics}_\text{S}$) in Table \ref{tab:texts}. For RNN-based PCAE, presented sentences intuitively match the given attribute well. For instance, in Yelp tense task, context assigned with ``Past'' attribute has signal words of ``was'' and ``expected''. In Titles $\text{topics}_\text{S}$ task, sentence that belong to different conditions owns keywords including the names of celebrities (``kardashia'', ``beyoncé''), name of cell phones (``iphone'') or code for diseases (``ebola virus''). For BART-based PCAE, sentences are more complex and fluent. For instance, in Yelp sentiment task, sentence with ``Positive'' label presents ``good'', ``smooth'' and ``tasty'' in the comment. As in Titles $\text{topics}_\text{S}$ task, sentences given different topics show their own features, such as text belongs to the ``Business'' topic talks about the company value of Tinder, while text from ``Health'' focuses on the calorie of food.
\begin{figure}[]
\centering
\includegraphics[width=1.0\linewidth]{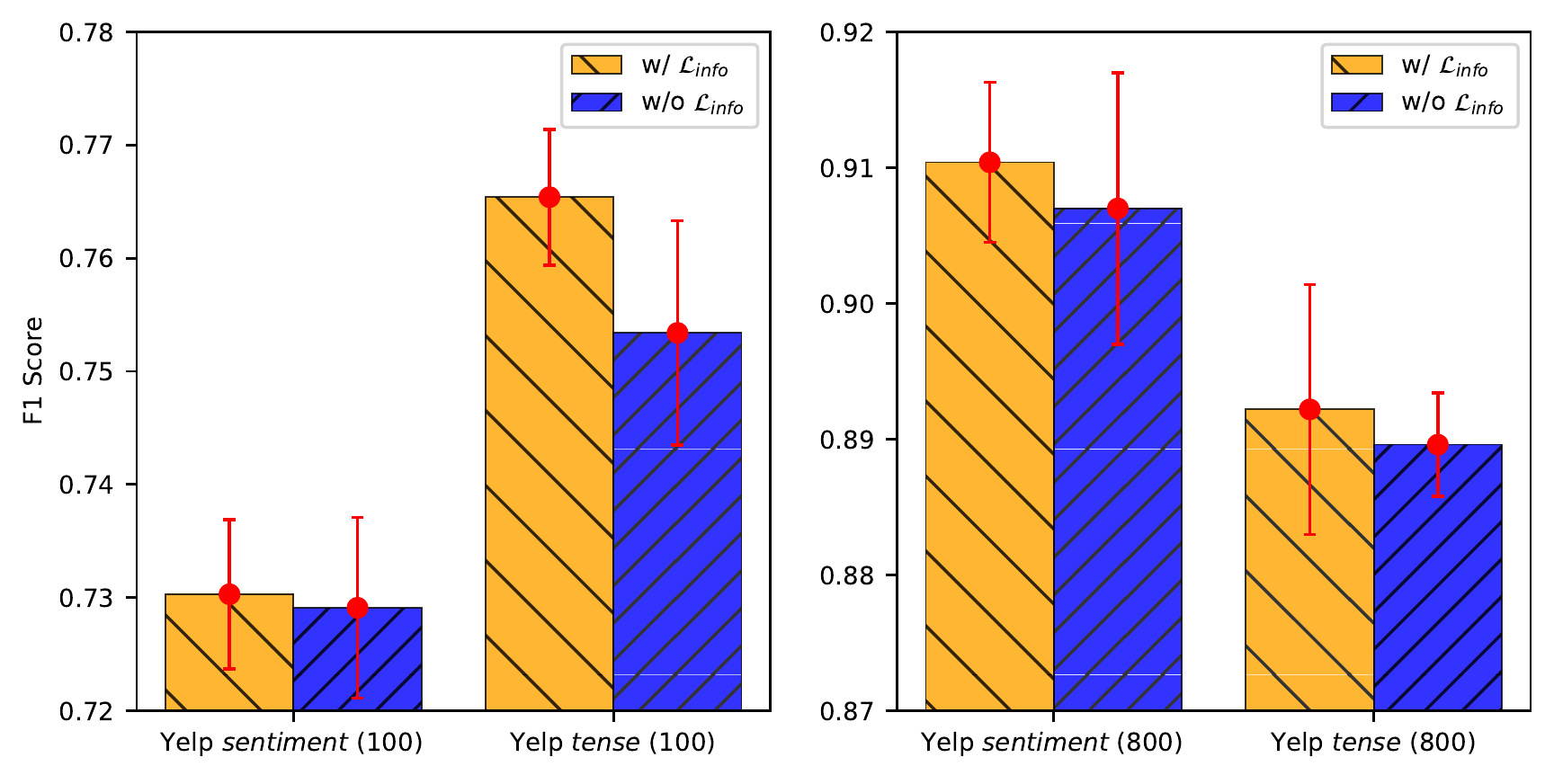}
\setlength{\abovecaptionskip}{-0.11cm}
\caption{Ablation study with F1 indicator of RNN-based PCAE with 12 layers in terms of latent regulator $\mathcal{L}_{\text{info}}$ on sentiment and tense tasks. Number in parentheses behind task name is the label number for each class for model training.}
\label{fig:latent_regulator_ablation}
\end{figure}
\subsubsection{Ablation Study}
We conducted all ablation experiments on RNN-based PCAE and similar behaviour should be observed for BART-based one. Firstly, we specifically analysis the impact of different broadcasting layer and labeled samples for training to the control of generated texts on five tasks. As shown in Figure \ref{fig:ablationln}, (1) More broadcasting layers can bring up the control ability of the proposed model until it is 12, that is 10 or 12 layers of \textit{Broadcasting Net} generally perform the best among all presented experiments results. (2) With the number of labels increases, the F1 scores climb higher in general. And keep increasing the number of labeled samples will finally make the our model in full supervision.

Secondly, we explore the effectiveness of latent regulator $\mathcal{L}_{\text{info}}$. Theoretically speaking, the latent regulator i.e., mutual information maximization term between code $\boldsymbol{z_l}$ and observed data $\boldsymbol{X}$ can bring the model to a deeper learning stage, thus helpful in enhancing the controllable ability of our model. To verify the impact of such latent regulator, we use test F1 score on sentiment and tense tasks with $100$ and $800$ labeled sample for each categories respectively, further train PluginAE with or without $\mathcal{L}_{\text{info}}$ and report the F1 score in Figure \ref{fig:latent_regulator_ablation}. From the results, we can see that (1) the latent regulator helps strengthen control capability of our model. (2) F1 results of models with $\mathcal{L}_{\text{info}}$ gain a narrower standard deviation than models without $\mathcal{L}_{\text{info}}$, which demonstrates that the latent regulator enhances the robustness of PCAE.


\vspace{-0.1mm}
\section{Conclusion}
In this work, we present a model-agnostic semi-supervised controllable text generation framework PCAE, which leads to remarkable empirical performance on both RNN-based and pre-trained BART-based settings. By adding \textit{Broadcasting Net} to existing plug-and-play system, we put this mainstay framework into a more universal pattern to generate controllable textual contents with auto-encoders. The visualization of learned local latent prior justifies the model optimization takes place in the hidden space. More importantly, with higher degree of controllability and competitive diversity of output texts plus less resource costs to apply the proposed framework, PCAE shows promising results on validating the effectiveness of proposed \textit{Broadcasting Net}. To make our model more practical in real-world scenarios, we apply both plain RNN trained from the scratch and pre-trained BART to our framework, and consistent results proves the effectiveness of the proposed framework. 
\section{Acknowledgment}
This work was supported in part by the National Natural Science Foundation of China (No.U1936208 and No.61862002, No.U1705261).

\bibliographystyle{cas-model2-names}

\bibliography{cas-refs}

\begin{thebibliography}{56}
\expandafter\ifx\csname natexlab\endcsname\relax\def\natexlab#1{#1}\fi
\providecommand{\url}[1]{\texttt{#1}}
\providecommand{\href}[2]{#2}
\providecommand{\path}[1]{#1}
\providecommand{\DOIprefix}{doi:}
\providecommand{\ArXivprefix}{arXiv:}
\providecommand{\URLprefix}{URL: }
\providecommand{\Pubmedprefix}{pmid:}
\providecommand{\doi}[1]{\href{http://dx.doi.org/#1}{\path{#1}}}
\providecommand{\Pubmed}[1]{\href{pmid:#1}{\path{#1}}}
\providecommand{\bibinfo}[2]{#2}
\ifx\xfnm\relax \def\xfnm[#1]{\unskip,\space#1}\fi
\bibitem[{Bottou(2010)}]{bottou2010large}
\bibinfo{author}{Bottou, L.}, \bibinfo{year}{2010}.
\newblock \bibinfo{title}{Large-scale machine learning with stochastic gradient
  descent}, in: \bibinfo{booktitle}{Proceedings of COMPSTAT'2010}.
  \bibinfo{publisher}{Springer}, pp. \bibinfo{pages}{177--186}.
\bibitem[{Bowman et~al.(2016)Bowman, Vilnis, Vinyals, Dai, Jozefowicz and
  Bengio}]{bowman2016generating}
\bibinfo{author}{Bowman, S.}, \bibinfo{author}{Vilnis, L.},
  \bibinfo{author}{Vinyals, O.}, \bibinfo{author}{Dai, A.},
  \bibinfo{author}{Jozefowicz, R.}, \bibinfo{author}{Bengio, S.},
  \bibinfo{year}{2016}.
\newblock \bibinfo{title}{Generating sentences from a continuous space}, in:
  \bibinfo{booktitle}{Proceedings of The 20th SIGNLL Conference on
  Computational Natural Language Learning}, pp. \bibinfo{pages}{10--21}.
\bibitem[{Chen et~al.(2016)Chen, Duan, Houthooft, Schulman, Sutskever and
  Abbeel}]{chen2016infogan}
\bibinfo{author}{Chen, X.}, \bibinfo{author}{Duan, Y.},
  \bibinfo{author}{Houthooft, R.}, \bibinfo{author}{Schulman, J.},
  \bibinfo{author}{Sutskever, I.}, \bibinfo{author}{Abbeel, P.},
  \bibinfo{year}{2016}.
\newblock \bibinfo{title}{Infogan: Interpretable representation learning by
  information maximizing generative adversarial nets}, in:
  \bibinfo{booktitle}{Proceedings of the 30th International Conference on
  Neural Information Processing Systems}, pp. \bibinfo{pages}{2180--2188}.
\bibitem[{Dathathri et~al.(2019)Dathathri, Madotto, Lan, Hung, Frank, Molino,
  Yosinski and Liu}]{dathathri2019plug}
\bibinfo{author}{Dathathri, S.}, \bibinfo{author}{Madotto, A.},
  \bibinfo{author}{Lan, J.}, \bibinfo{author}{Hung, J.},
  \bibinfo{author}{Frank, E.}, \bibinfo{author}{Molino, P.},
  \bibinfo{author}{Yosinski, J.}, \bibinfo{author}{Liu, R.},
  \bibinfo{year}{2019}.
\newblock \bibinfo{title}{Plug and play language models: A simple approach to
  controlled text generation}.
\newblock \bibinfo{journal}{arXiv preprint arXiv:1912.02164} .
\bibitem[{Devlin et~al.(2019)Devlin, Chang, Lee and Toutanova}]{devlin2019bert}
\bibinfo{author}{Devlin, J.}, \bibinfo{author}{Chang, M.W.},
  \bibinfo{author}{Lee, K.}, \bibinfo{author}{Toutanova, K.},
  \bibinfo{year}{2019}.
\newblock \bibinfo{title}{Bert: Pre-training of deep bidirectional transformers
  for language understanding}, in: \bibinfo{booktitle}{Proceedings of the 2019
  Conference of the North American Chapter of the Association for Computational
  Linguistics: Human Language Technologies, Volume 1 (Long and Short Papers)},
  pp. \bibinfo{pages}{4171--4186}.
\bibitem[{Duan et~al.(2020)Duan, Xu, Pei, Han and Li}]{duan2020pre}
\bibinfo{author}{Duan, Y.}, \bibinfo{author}{Xu, C.}, \bibinfo{author}{Pei,
  J.}, \bibinfo{author}{Han, J.}, \bibinfo{author}{Li, C.},
  \bibinfo{year}{2020}.
\newblock \bibinfo{title}{Pre-train and plug-in: Flexible conditional text
  generation with variational auto-encoders}, in:
  \bibinfo{booktitle}{Proceedings of the 58th Annual Meeting of the Association
  for Computational Linguistics}, pp. \bibinfo{pages}{253--262}.
\bibitem[{Elhadad(1990)}]{elhadad1990constraint}
\bibinfo{author}{Elhadad, M.}, \bibinfo{year}{1990}.
\newblock \bibinfo{title}{Constraint-based text generation using local
  constraints and argumentation to generate a turn in conversation} .
\bibitem[{Fang et~al.(2019)Fang, Li, Gao, Dong and Chen}]{fang2019implicit}
\bibinfo{author}{Fang, L.}, \bibinfo{author}{Li, C.}, \bibinfo{author}{Gao,
  J.}, \bibinfo{author}{Dong, W.}, \bibinfo{author}{Chen, C.},
  \bibinfo{year}{2019}.
\newblock \bibinfo{title}{Implicit deep latent variable models for text
  generation}, in: \bibinfo{booktitle}{Proceedings of the 2019 Conference on
  Empirical Methods in Natural Language Processing and the 9th International
  Joint Conference on Natural Language Processing (EMNLP-IJCNLP)}, pp.
  \bibinfo{pages}{3946--3956}.
\bibitem[{Fang et~al.(2021)Fang, Zeng, Liu, Bo, Dong and
  Chen}]{fang2021transformer}
\bibinfo{author}{Fang, L.}, \bibinfo{author}{Zeng, T.}, \bibinfo{author}{Liu,
  C.}, \bibinfo{author}{Bo, L.}, \bibinfo{author}{Dong, W.},
  \bibinfo{author}{Chen, C.}, \bibinfo{year}{2021}.
\newblock \bibinfo{title}{Transformer-based conditional variational autoencoder
  for controllable story generation}.
\newblock \bibinfo{journal}{arXiv preprint arXiv:2101.00828} .
\bibitem[{Ficler and Goldberg(2017)}]{ficler2017controlling}
\bibinfo{author}{Ficler, J.}, \bibinfo{author}{Goldberg, Y.},
  \bibinfo{year}{2017}.
\newblock \bibinfo{title}{Controlling linguistic style aspects in neural
  language generation}.
\newblock \bibinfo{journal}{EMNLP 2017} , \bibinfo{pages}{94}.
\bibitem[{Fu et~al.(2019)Fu, Li, Liu, Gao, Celikyilmaz and
  Carin}]{fu2019cyclical}
\bibinfo{author}{Fu, H.}, \bibinfo{author}{Li, C.}, \bibinfo{author}{Liu, X.},
  \bibinfo{author}{Gao, J.}, \bibinfo{author}{Celikyilmaz, A.},
  \bibinfo{author}{Carin, L.}, \bibinfo{year}{2019}.
\newblock \bibinfo{title}{Cyclical annealing schedule: A simple approach to
  mitigating kl vanishing}, in: \bibinfo{booktitle}{Proceedings of the 2019
  Conference of the North American Chapter of the Association for Computational
  Linguistics: Human Language Technologies, Volume 1 (Long and Short Papers)},
  pp. \bibinfo{pages}{240--250}.
\bibitem[{Fu et~al.(2018)Fu, Tan, Peng, Zhao and Yan}]{fu2018style}
\bibinfo{author}{Fu, Z.}, \bibinfo{author}{Tan, X.}, \bibinfo{author}{Peng,
  N.}, \bibinfo{author}{Zhao, D.}, \bibinfo{author}{Yan, R.},
  \bibinfo{year}{2018}.
\newblock \bibinfo{title}{Style transfer in text: Exploration and evaluation},
  in: \bibinfo{booktitle}{Proceedings of the AAAI Conference on Artificial
  Intelligence}.
\bibitem[{Garbacea and Mei(2020)}]{garbacea2020neural}
\bibinfo{author}{Garbacea, C.}, \bibinfo{author}{Mei, Q.},
  \bibinfo{year}{2020}.
\newblock \bibinfo{title}{Neural language generation: Formulation, methods, and
  evaluation}.
\newblock \bibinfo{journal}{arXiv preprint arXiv:2007.15780} .
\bibitem[{Gretton et~al.(2006)Gretton, Borgwardt, Rasch, Sch{\"o}lkopf and
  Smola}]{gretton2006kernel}
\bibinfo{author}{Gretton, A.}, \bibinfo{author}{Borgwardt, K.},
  \bibinfo{author}{Rasch, M.}, \bibinfo{author}{Sch{\"o}lkopf, B.},
  \bibinfo{author}{Smola, A.}, \bibinfo{year}{2006}.
\newblock \bibinfo{title}{A kernel method for the two-sample-problem}.
\newblock \bibinfo{journal}{Advances in neural information processing systems}
  \bibinfo{volume}{19}, \bibinfo{pages}{513--520}.
\bibitem[{He et~al.(2021)He, Zhou, Ma, Berg-Kirkpatrick and
  Neubig}]{he2021towards}
\bibinfo{author}{He, J.}, \bibinfo{author}{Zhou, C.}, \bibinfo{author}{Ma, X.},
  \bibinfo{author}{Berg-Kirkpatrick, T.}, \bibinfo{author}{Neubig, G.},
  \bibinfo{year}{2021}.
\newblock \bibinfo{title}{Towards a unified view of parameter-efficient
  transfer learning}, in: \bibinfo{booktitle}{International Conference on
  Learning Representations}.
\bibitem[{He et~al.(2016)He, Zhang, Ren and Sun}]{he2016deep}
\bibinfo{author}{He, K.}, \bibinfo{author}{Zhang, X.}, \bibinfo{author}{Ren,
  S.}, \bibinfo{author}{Sun, J.}, \bibinfo{year}{2016}.
\newblock \bibinfo{title}{Deep residual learning for image recognition}, in:
  \bibinfo{booktitle}{Proceedings of the IEEE conference on computer vision and
  pattern recognition}, pp. \bibinfo{pages}{770--778}.
\bibitem[{Hochreiter and Schmidhuber(1997)}]{hochreiter1997long}
\bibinfo{author}{Hochreiter, S.}, \bibinfo{author}{Schmidhuber, J.},
  \bibinfo{year}{1997}.
\newblock \bibinfo{title}{Long short-term memory}.
\newblock \bibinfo{journal}{Neural computation} \bibinfo{volume}{9},
  \bibinfo{pages}{1735--1780}.
\bibitem[{Holtzman et~al.(2019)Holtzman, Buys, Du, Forbes and
  Choi}]{holtzman2019curious}
\bibinfo{author}{Holtzman, A.}, \bibinfo{author}{Buys, J.},
  \bibinfo{author}{Du, L.}, \bibinfo{author}{Forbes, M.},
  \bibinfo{author}{Choi, Y.}, \bibinfo{year}{2019}.
\newblock \bibinfo{title}{The curious case of neural text degeneration}, in:
  \bibinfo{booktitle}{International Conference on Learning Representations}.
\bibitem[{Houlsby et~al.(2019)Houlsby, Giurgiu, Jastrzebski, Morrone,
  De~Laroussilhe, Gesmundo, Attariyan and Gelly}]{houlsby2019parameter}
\bibinfo{author}{Houlsby, N.}, \bibinfo{author}{Giurgiu, A.},
  \bibinfo{author}{Jastrzebski, S.}, \bibinfo{author}{Morrone, B.},
  \bibinfo{author}{De~Laroussilhe, Q.}, \bibinfo{author}{Gesmundo, A.},
  \bibinfo{author}{Attariyan, M.}, \bibinfo{author}{Gelly, S.},
  \bibinfo{year}{2019}.
\newblock \bibinfo{title}{Parameter-efficient transfer learning for nlp}, in:
  \bibinfo{booktitle}{International Conference on Machine Learning},
  \bibinfo{organization}{PMLR}. pp. \bibinfo{pages}{2790--2799}.
\bibitem[{Hu et~al.(2017)Hu, Yang, Liang, Salakhutdinov and
  Xing}]{hu2017toward}
\bibinfo{author}{Hu, Z.}, \bibinfo{author}{Yang, Z.}, \bibinfo{author}{Liang,
  X.}, \bibinfo{author}{Salakhutdinov, R.}, \bibinfo{author}{Xing, E.P.},
  \bibinfo{year}{2017}.
\newblock \bibinfo{title}{Toward controlled generation of text}, in:
  \bibinfo{booktitle}{International Conference on Machine Learning},
  \bibinfo{organization}{PMLR}. pp. \bibinfo{pages}{1587--1596}.
\bibitem[{Huang et~al.(2017)Huang, Liu, Van Der~Maaten and
  Weinberger}]{huang2017densely}
\bibinfo{author}{Huang, G.}, \bibinfo{author}{Liu, Z.}, \bibinfo{author}{Van
  Der~Maaten, L.}, \bibinfo{author}{Weinberger, K.Q.}, \bibinfo{year}{2017}.
\newblock \bibinfo{title}{Densely connected convolutional networks}, in:
  \bibinfo{booktitle}{Proceedings of the IEEE conference on computer vision and
  pattern recognition}, pp. \bibinfo{pages}{4700--4708}.
\bibitem[{Keskar et~al.(2019)Keskar, McCann, Varshney, Xiong and
  Socher}]{keskar2019ctrl}
\bibinfo{author}{Keskar, N.S.}, \bibinfo{author}{McCann, B.},
  \bibinfo{author}{Varshney, L.R.}, \bibinfo{author}{Xiong, C.},
  \bibinfo{author}{Socher, R.}, \bibinfo{year}{2019}.
\newblock \bibinfo{title}{Ctrl: A conditional transformer language model for
  controllable generation}.
\newblock \bibinfo{journal}{arXiv preprint arXiv:1909.05858} .
\bibitem[{Kingma and Ba(2014)}]{kingma2014adam}
\bibinfo{author}{Kingma, D.P.}, \bibinfo{author}{Ba, J.}, \bibinfo{year}{2014}.
\newblock \bibinfo{title}{Adam: A method for stochastic optimization}.
\newblock \bibinfo{journal}{arXiv preprint arXiv:1412.6980} .
\bibitem[{Kingma et~al.(2014)Kingma, Mohamed, Rezende and
  Welling}]{kingma2014semi}
\bibinfo{author}{Kingma, D.P.}, \bibinfo{author}{Mohamed, S.},
  \bibinfo{author}{Rezende, D.J.}, \bibinfo{author}{Welling, M.},
  \bibinfo{year}{2014}.
\newblock \bibinfo{title}{Semi-supervised learning with deep generative
  models}, in: \bibinfo{booktitle}{Advances in neural information processing
  systems}, pp. \bibinfo{pages}{3581--3589}.
\bibitem[{Krause et~al.(2021)Krause, Gotmare, McCann, Keskar, Joty, Socher and
  Rajani}]{krause2021gedi}
\bibinfo{author}{Krause, B.}, \bibinfo{author}{Gotmare, A.D.},
  \bibinfo{author}{McCann, B.}, \bibinfo{author}{Keskar, N.S.},
  \bibinfo{author}{Joty, S.}, \bibinfo{author}{Socher, R.},
  \bibinfo{author}{Rajani, N.F.}, \bibinfo{year}{2021}.
\newblock \bibinfo{title}{Gedi: Generative discriminator guided sequence
  generation}, in: \bibinfo{booktitle}{Findings of the Association for
  Computational Linguistics: EMNLP 2021}, pp. \bibinfo{pages}{4929--4952}.
\bibitem[{Lewis et~al.(2020)Lewis, Liu, Goyal, Ghazvininejad, Mohamed, Levy,
  Stoyanov and Zettlemoyer}]{lewis2020bart}
\bibinfo{author}{Lewis, M.}, \bibinfo{author}{Liu, Y.}, \bibinfo{author}{Goyal,
  N.}, \bibinfo{author}{Ghazvininejad, M.}, \bibinfo{author}{Mohamed, A.},
  \bibinfo{author}{Levy, O.}, \bibinfo{author}{Stoyanov, V.},
  \bibinfo{author}{Zettlemoyer, L.}, \bibinfo{year}{2020}.
\newblock \bibinfo{title}{Bart: Denoising sequence-to-sequence pre-training for
  natural language generation, translation, and comprehension}, in:
  \bibinfo{booktitle}{Proceedings of the 58th Annual Meeting of the Association
  for Computational Linguistics}, pp. \bibinfo{pages}{7871--7880}.
\bibitem[{Li et~al.(2019)Li, He, Neubig, Berg-Kirkpatrick and
  Yang}]{li2019surprisingly}
\bibinfo{author}{Li, B.}, \bibinfo{author}{He, J.}, \bibinfo{author}{Neubig,
  G.}, \bibinfo{author}{Berg-Kirkpatrick, T.}, \bibinfo{author}{Yang, Y.},
  \bibinfo{year}{2019}.
\newblock \bibinfo{title}{A surprisingly effective fix for deep latent variable
  modeling of text}, in: \bibinfo{booktitle}{Proceedings of the 2019 Conference
  on Empirical Methods in Natural Language Processing and the 9th International
  Joint Conference on Natural Language Processing (EMNLP-IJCNLP)}, pp.
  \bibinfo{pages}{3603--3614}.
\bibitem[{Li et~al.(2020)Li, Gao, Li, Peng, Li, Zhang and Gao}]{li2020optimus}
\bibinfo{author}{Li, C.}, \bibinfo{author}{Gao, X.}, \bibinfo{author}{Li, Y.},
  \bibinfo{author}{Peng, B.}, \bibinfo{author}{Li, X.}, \bibinfo{author}{Zhang,
  Y.}, \bibinfo{author}{Gao, J.}, \bibinfo{year}{2020}.
\newblock \bibinfo{title}{Optimus: Organizing sentences via pre-trained
  modeling of a latent space}, in: \bibinfo{booktitle}{Proceedings of the 2020
  Conference on Empirical Methods in Natural Language Processing (EMNLP)}, pp.
  \bibinfo{pages}{4678--4699}.
\bibitem[{Li et~al.(2016)Li, Galley, Brockett, Gao and Dolan}]{li2016diversity}
\bibinfo{author}{Li, J.}, \bibinfo{author}{Galley, M.},
  \bibinfo{author}{Brockett, C.}, \bibinfo{author}{Gao, J.},
  \bibinfo{author}{Dolan, W.B.}, \bibinfo{year}{2016}.
\newblock \bibinfo{title}{A diversity-promoting objective function for neural
  conversation models}, in: \bibinfo{booktitle}{Proceedings of the 2016
  Conference of the North American Chapter of the Association for Computational
  Linguistics: Human Language Technologies}, pp. \bibinfo{pages}{110--119}.
\bibitem[{Li and Liang(2021)}]{li2021prefix}
\bibinfo{author}{Li, X.L.}, \bibinfo{author}{Liang, P.}, \bibinfo{year}{2021}.
\newblock \bibinfo{title}{Prefix-tuning: Optimizing continuous prompts for
  generation}, in: \bibinfo{booktitle}{Proceedings of the 59th Annual Meeting
  of the Association for Computational Linguistics and the 11th International
  Joint Conference on Natural Language Processing (Volume 1: Long Papers)}, pp.
  \bibinfo{pages}{4582--4597}.
\bibitem[{Li et~al.(2015)Li, Swersky and Zemel}]{li2015generative}
\bibinfo{author}{Li, Y.}, \bibinfo{author}{Swersky, K.},
  \bibinfo{author}{Zemel, R.}, \bibinfo{year}{2015}.
\newblock \bibinfo{title}{Generative moment matching networks}, in:
  \bibinfo{booktitle}{International Conference on Machine Learning},
  \bibinfo{organization}{PMLR}. pp. \bibinfo{pages}{1718--1727}.
\bibitem[{Liu and Liu(2019)}]{liu2019transformer}
\bibinfo{author}{Liu, D.}, \bibinfo{author}{Liu, G.}, \bibinfo{year}{2019}.
\newblock \bibinfo{title}{A transformer-based variational autoencoder for
  sentence generation}, in: \bibinfo{booktitle}{2019 International Joint
  Conference on Neural Networks (IJCNN)}, \bibinfo{organization}{IEEE}. pp.
  \bibinfo{pages}{1--7}.
\bibitem[{Loshchilov and Hutter(2018)}]{loshchilov2018decoupled}
\bibinfo{author}{Loshchilov, I.}, \bibinfo{author}{Hutter, F.},
  \bibinfo{year}{2018}.
\newblock \bibinfo{title}{Decoupled weight decay regularization}, in:
  \bibinfo{booktitle}{International Conference on Learning Representations}.
\bibitem[{Van~der Maaten and Hinton(2008)}]{van2008visualizing}
\bibinfo{author}{Van~der Maaten, L.}, \bibinfo{author}{Hinton, G.},
  \bibinfo{year}{2008}.
\newblock \bibinfo{title}{Visualizing data using t-sne.}
\newblock \bibinfo{journal}{Journal of machine learning research}
  \bibinfo{volume}{9}.
\bibitem[{Mai et~al.(2020)Mai, Pappas, Montero, Smith and
  Henderson}]{mai2020plug}
\bibinfo{author}{Mai, F.}, \bibinfo{author}{Pappas, N.},
  \bibinfo{author}{Montero, I.}, \bibinfo{author}{Smith, N.A.},
  \bibinfo{author}{Henderson, J.}, \bibinfo{year}{2020}.
\newblock \bibinfo{title}{Plug and play autoencoders for conditional text
  generation}.
\newblock \bibinfo{journal}{arXiv preprint arXiv:2010.02983} .
\bibitem[{Makhzani et~al.(2015)Makhzani, Shlens, Jaitly, Goodfellow and
  Frey}]{makhzani2015adversarial}
\bibinfo{author}{Makhzani, A.}, \bibinfo{author}{Shlens, J.},
  \bibinfo{author}{Jaitly, N.}, \bibinfo{author}{Goodfellow, I.},
  \bibinfo{author}{Frey, B.}, \bibinfo{year}{2015}.
\newblock \bibinfo{title}{Adversarial autoencoders}.
\newblock \bibinfo{journal}{arXiv preprint arXiv:1511.05644} .
\bibitem[{Meehan(1977)}]{meehan1977tale}
\bibinfo{author}{Meehan, J.R.}, \bibinfo{year}{1977}.
\newblock \bibinfo{title}{Tale-spin, an interactive program that writes
  stories.}, in: \bibinfo{booktitle}{Ijcai}, p. \bibinfo{pages}{9198}.
\bibitem[{Park and Lee(2021)}]{park2021finetuning}
\bibinfo{author}{Park, S.}, \bibinfo{author}{Lee, J.}, \bibinfo{year}{2021}.
\newblock \bibinfo{title}{Finetuning pretrained transformers into variational
  autoencoders}, in: \bibinfo{booktitle}{Proceedings of the Second Workshop on
  Insights from Negative Results in NLP}, pp. \bibinfo{pages}{29--35}.
\bibitem[{Pascual et~al.(2020)Pascual, Egressy, Bolli and
  Wattenhofer}]{pascual2020directed}
\bibinfo{author}{Pascual, D.}, \bibinfo{author}{Egressy, B.},
  \bibinfo{author}{Bolli, F.}, \bibinfo{author}{Wattenhofer, R.},
  \bibinfo{year}{2020}.
\newblock \bibinfo{title}{Directed beam search: Plug-and-play lexically
  constrained language generation}.
\newblock \bibinfo{journal}{arXiv preprint arXiv:2012.15416} .
\bibitem[{Pelsmaeker and Aziz(2020)}]{pelsmaeker2020effective}
\bibinfo{author}{Pelsmaeker, T.}, \bibinfo{author}{Aziz, W.},
  \bibinfo{year}{2020}.
\newblock \bibinfo{title}{Effective estimation of deep generative language
  models}, in: \bibinfo{booktitle}{Proceedings of the 58th Annual Meeting of
  the Association for Computational Linguistics}, pp.
  \bibinfo{pages}{7220--7236}.
\bibitem[{Radford et~al.(2019)Radford, Wu, Child, Luan, Amodei, Sutskever
  et~al.}]{radford2019language}
\bibinfo{author}{Radford, A.}, \bibinfo{author}{Wu, J.},
  \bibinfo{author}{Child, R.}, \bibinfo{author}{Luan, D.},
  \bibinfo{author}{Amodei, D.}, \bibinfo{author}{Sutskever, I.}, et~al.,
  \bibinfo{year}{2019}.
\newblock \bibinfo{title}{Language models are unsupervised multitask learners}.
\newblock \bibinfo{journal}{OpenAI blog} \bibinfo{volume}{1},
  \bibinfo{pages}{9}.
\bibitem[{Raffel et~al.(2020)Raffel, Shazeer, Roberts, Lee, Narang, Matena,
  Zhou, Li, Liu et~al.}]{raffel2020exploring}
\bibinfo{author}{Raffel, C.}, \bibinfo{author}{Shazeer, N.},
  \bibinfo{author}{Roberts, A.}, \bibinfo{author}{Lee, K.},
  \bibinfo{author}{Narang, S.}, \bibinfo{author}{Matena, M.},
  \bibinfo{author}{Zhou, Y.}, \bibinfo{author}{Li, W.}, \bibinfo{author}{Liu,
  P.J.}, et~al., \bibinfo{year}{2020}.
\newblock \bibinfo{title}{Exploring the limits of transfer learning with a
  unified text-to-text transformer.}
\newblock \bibinfo{journal}{J. Mach. Learn. Res.} \bibinfo{volume}{21},
  \bibinfo{pages}{1--67}.
\bibitem[{Razavi et~al.(2019)Razavi, van~den Oord and
  Vinyals}]{razavi2019generating}
\bibinfo{author}{Razavi, A.}, \bibinfo{author}{van~den Oord, A.},
  \bibinfo{author}{Vinyals, O.}, \bibinfo{year}{2019}.
\newblock \bibinfo{title}{Generating diverse high-fidelity images with
  vq-vae-2}, in: \bibinfo{booktitle}{Advances in neural information processing
  systems}, pp. \bibinfo{pages}{14866--14876}.
\bibitem[{Rezaabad and Vishwanath(2020)}]{rezaabad2020learning}
\bibinfo{author}{Rezaabad, A.L.}, \bibinfo{author}{Vishwanath, S.},
  \bibinfo{year}{2020}.
\newblock \bibinfo{title}{Learning representations by maximizing mutual
  information in variational autoencoders}, in: \bibinfo{booktitle}{2020 IEEE
  International Symposium on Information Theory (ISIT)},
  \bibinfo{organization}{IEEE}. pp. \bibinfo{pages}{2729--2734}.
\bibitem[{Shen et~al.(2017)Shen, Lei, Barzilay and Jaakkola}]{shen2017style}
\bibinfo{author}{Shen, T.}, \bibinfo{author}{Lei, T.},
  \bibinfo{author}{Barzilay, R.}, \bibinfo{author}{Jaakkola, T.},
  \bibinfo{year}{2017}.
\newblock \bibinfo{title}{Style transfer from non-parallel text by
  cross-alignment}.
\newblock \bibinfo{journal}{arXiv preprint arXiv:1705.09655} .
\bibitem[{Shen et~al.(2020)Shen, Mueller, Barzilay and
  Jaakkola}]{shen2020educating}
\bibinfo{author}{Shen, T.}, \bibinfo{author}{Mueller, J.},
  \bibinfo{author}{Barzilay, R.}, \bibinfo{author}{Jaakkola, T.},
  \bibinfo{year}{2020}.
\newblock \bibinfo{title}{Educating text autoencoders: Latent representation
  guidance via denoising}, in: \bibinfo{booktitle}{International Conference on
  Machine Learning}, \bibinfo{organization}{PMLR}. pp.
  \bibinfo{pages}{8719--8729}.
\bibitem[{Tang et~al.(2019)Tang, Li and Jin}]{tang2019topic}
\bibinfo{author}{Tang, H.}, \bibinfo{author}{Li, M.}, \bibinfo{author}{Jin,
  B.}, \bibinfo{year}{2019}.
\newblock \bibinfo{title}{A topic augmented text generation model: Joint
  learning of semantics and structural features}, in:
  \bibinfo{booktitle}{Proceedings of the 2019 Conference on Empirical Methods
  in Natural Language Processing and the 9th International Joint Conference on
  Natural Language Processing (EMNLP-IJCNLP)}, pp. \bibinfo{pages}{5090--5099}.
\bibitem[{Tu et~al.(2022)Tu, Yang, Yang, Zhang and Huang}]{tu2022adavae}
\bibinfo{author}{Tu, H.}, \bibinfo{author}{Yang, Z.}, \bibinfo{author}{Yang,
  J.}, \bibinfo{author}{Zhang, S.}, \bibinfo{author}{Huang, Y.},
  \bibinfo{year}{2022}.
\newblock \bibinfo{title}{Adavae: Exploring adaptive gpt-2s in variational
  auto-encoders for language modeling}.
\newblock \bibinfo{journal}{arXiv preprint arXiv:2205.05862} .
\bibitem[{Vincent et~al.(2010)Vincent, Larochelle, Lajoie, Bengio, Manzagol and
  Bottou}]{vincent2010stacked}
\bibinfo{author}{Vincent, P.}, \bibinfo{author}{Larochelle, H.},
  \bibinfo{author}{Lajoie, I.}, \bibinfo{author}{Bengio, Y.},
  \bibinfo{author}{Manzagol, P.A.}, \bibinfo{author}{Bottou, L.},
  \bibinfo{year}{2010}.
\newblock \bibinfo{title}{Stacked denoising autoencoders: Learning useful
  representations in a deep network with a local denoising criterion.}
\newblock \bibinfo{journal}{Journal of machine learning research}
  \bibinfo{volume}{11}.
\bibitem[{Wallace et~al.(2019)Wallace, Feng, Kandpal, Gardner and
  Singh}]{wallace2019universal}
\bibinfo{author}{Wallace, E.}, \bibinfo{author}{Feng, S.},
  \bibinfo{author}{Kandpal, N.}, \bibinfo{author}{Gardner, M.},
  \bibinfo{author}{Singh, S.}, \bibinfo{year}{2019}.
\newblock \bibinfo{title}{Universal adversarial triggers for attacking and
  analyzing nlp}, in: \bibinfo{booktitle}{Proceedings of the 2019 Conference on
  Empirical Methods in Natural Language Processing and the 9th International
  Joint Conference on Natural Language Processing (EMNLP-IJCNLP)}, pp.
  \bibinfo{pages}{2153--2162}.
\bibitem[{Wang and Wan(2018)}]{wang2018sentigan}
\bibinfo{author}{Wang, K.}, \bibinfo{author}{Wan, X.}, \bibinfo{year}{2018}.
\newblock \bibinfo{title}{Sentigan: Generating sentimental texts via mixture
  adversarial networks.}, in: \bibinfo{booktitle}{IJCAI}, pp.
  \bibinfo{pages}{4446--4452}.
\bibitem[{Wang et~al.(2017)Wang, Schwing and Lazebnik}]{wang2017diverse}
\bibinfo{author}{Wang, L.}, \bibinfo{author}{Schwing, A.G.},
  \bibinfo{author}{Lazebnik, S.}, \bibinfo{year}{2017}.
\newblock \bibinfo{title}{Diverse and accurate image description using a
  variational auto-encoder with an additive gaussian encoding space}, in:
  \bibinfo{booktitle}{NIPS}.
\bibitem[{Wang et~al.(2019)Wang, Gan, Xu, Zhang, Wang, Shen, Chen and
  Carin}]{wang2019topic}
\bibinfo{author}{Wang, W.}, \bibinfo{author}{Gan, Z.}, \bibinfo{author}{Xu,
  H.}, \bibinfo{author}{Zhang, R.}, \bibinfo{author}{Wang, G.},
  \bibinfo{author}{Shen, D.}, \bibinfo{author}{Chen, C.},
  \bibinfo{author}{Carin, L.}, \bibinfo{year}{2019}.
\newblock \bibinfo{title}{Topic-guided variational auto-encoder for text
  generation}, in: \bibinfo{booktitle}{Proceedings of the 2019 Conference of
  the North American Chapter of the Association for Computational Linguistics:
  Human Language Technologies, Volume 1 (Long and Short Papers)}, pp.
  \bibinfo{pages}{166--177}.
\bibitem[{Xu et~al.(2020)Xu, Cheung and Cao}]{xu2020variational}
\bibinfo{author}{Xu, P.}, \bibinfo{author}{Cheung, J.C.K.},
  \bibinfo{author}{Cao, Y.}, \bibinfo{year}{2020}.
\newblock \bibinfo{title}{On variational learning of controllable
  representations for text without supervision}, in:
  \bibinfo{booktitle}{International Conference on Machine Learning},
  \bibinfo{organization}{PMLR}. pp. \bibinfo{pages}{10534--10543}.
\bibitem[{Yang et~al.(2017)Yang, Hu, Salakhutdinov and
  Berg-Kirkpatrick}]{yang2017improved}
\bibinfo{author}{Yang, Z.}, \bibinfo{author}{Hu, Z.},
  \bibinfo{author}{Salakhutdinov, R.}, \bibinfo{author}{Berg-Kirkpatrick, T.},
  \bibinfo{year}{2017}.
\newblock \bibinfo{title}{Improved variational autoencoders for text modeling
  using dilated convolutions}, in: \bibinfo{booktitle}{International conference
  on machine learning}, \bibinfo{organization}{PMLR}. pp.
  \bibinfo{pages}{3881--3890}.
\bibitem[{Zhao et~al.(2017)Zhao, Song and Ermon}]{zhao2017infovae}
\bibinfo{author}{Zhao, S.}, \bibinfo{author}{Song, J.}, \bibinfo{author}{Ermon,
  S.}, \bibinfo{year}{2017}.
\newblock \bibinfo{title}{Infovae: Information maximizing variational
  autoencoders}.
\newblock \bibinfo{journal}{arXiv preprint arXiv:1706.02262} .

\end{thebibliography}





\end{document}